\documentclass[lettersize,journal]{IEEEtran}
\usepackage{amsmath,amsfonts}
\usepackage{algorithm}
\usepackage{array}
\usepackage[caption=false,font=normalsize,labelfont=sf,textfont=sf]{subfig}
\usepackage{textcomp}
\usepackage{stfloats}
\usepackage{url}
\usepackage{verbatim}
\usepackage{graphicx}
\usepackage{cite}
\usepackage{amssymb}  
\usepackage{xcolor}   
\usepackage{colortbl} 
\usepackage{booktabs}
\usepackage{algpseudocode}  
\usepackage{enumitem}
\usepackage{etoolbox}
\usepackage{multirow}
\algrenewcommand\algorithmiccomment[1]{\hfill\textcolor{gray}{// #1}}

\newcommand{\cmark}{\checkmark}

\newif\ifmain
\mainfalse

\begin{document}

\title{L2D2-GS: Learning to Densify for Feedforward Dynamic Gaussian Scene Reconstruction}

\author{
	Zetian~Song,
    Chenming~Wu,
    Junnan~Liu,
    Chitian~Sun,
    Liangliang~He,
    Hangjun~Ye,
	Jiaqi~Zhang,
	Siwei~Ma, ~\IEEEmembership{Fellow,~IEEE,}
	Wen~Gao, ~\IEEEmembership{Fellow,~IEEE,}\\	
        \thanks{Zetian~Song, Jiaqi~Zhang, Siwei~Ma and Wen Gao are with the State Key Laboratory of Multimedia Information Processing, School of Computer Science, Peking University, Beijing, 100871, China~(e-mail: songzt@pku.edu.cn, jqzhang@pku.edu.cn, swma@pku.edu.cn, wgao@pku.edu.cn). }
        \thanks{Zetian~Song, Chenming~Wu, Junnan~Liu, Chitian~Sun, Liangliang~He, Hangjun~Ye is with Xiaomi EV, Beijing, 100085, China~(email: songzt@pku.edu.cn, wcm15@mails.tsinghua.edu.cn, liujunnan5@xiaomi.com, sunchitian@xiaomi.com, heliangliang@xiaomi.com, yehangjun@xiaomi.com).}
	}
    

\markboth{Journal of \LaTeX\ Class Files,~Vol.~14, No.~8, August~2021}%
{Shell \MakeLowercase{\textit{et al.}}: A Sample Article Using IEEEtran.cls for IEEE Journals}



\maketitle

\begin{abstract}
High-fidelity reconstruction of dynamic urban environments is a cornerstone of autonomous driving simulation and large-scale world modeling. While 3D Gaussian Splatting (3DGS) has established a new standard for real-time rendering, its reliance on expensive per-scene optimization limits scalability. Conversely, recent feedforward methods that infer Gaussian parameters offer faster speed but face fundamental bottlenecks: they are memory-prohibitive at high resolutions and struggle to fuse dense multi-view observations consistently. This paper presents L2D2-GS, a unified framework that reformulates generalizable reconstruction not as a one-shot regression, but as a robust iterative process of optimization and densification. To resolve the ambiguity of supervision in primitive generation, we propose a self-supervised densification policy that derives explicit reward signals from global reconstruction gains to guide local densification. Furthermore, we mitigate irreversible early-stage artifacts through a geometric regularization mechanism, utilizing reparameterization to constrain the optimization manifold and prevent convergence to poor local optima. 
Extensive experiments on the PandaSet and Waymo datasets demonstrate that our method achieves state-of-the-art reconstruction fidelity and strong zero-shot generalization, while using fewer primitives than competing baselines.
\end{abstract}

\begin{IEEEkeywords}
Article submission, IEEE, IEEEtran, journal, \LaTeX, paper, template, typesetting.
\end{IEEEkeywords}

\begin{figure*}
\centering
\includegraphics[width=0.9\linewidth]{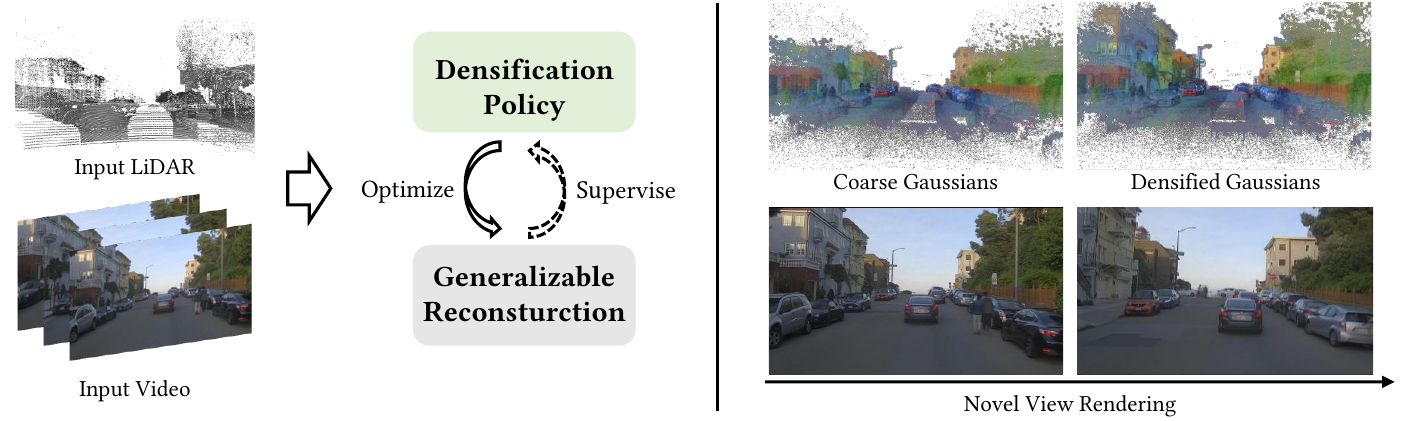}
\caption{Our proposed learning-to-densitfy framework enables generalizable reconstruction of dynamic urban scenes through a combination of feedforward prediction and iterative optimization with adaptive denstification.
To resolve supervision ambiguity in densification, we introduce a policy-based strategy that derives explicit rewards from reconstruction gains, enabling fully self-supervised learning.}
\vspace{-10pt}
\label{fig:teaser}
\end{figure*}


\maketitle

\section{Introduction}
\label{sec:intro}

\IEEEPARstart{R}{econstructing}
and synthesizing photorealistic views of dynamic urban environments is a cornerstone of autonomous driving simulation and large-scale world modeling. Although 3D Gaussian Splatting (3DGS)~\cite{kerbl20233d} has emerged as a powerful framework for high-fidelity reconstruction and real-time rendering \cite{yan2024street, chen2024omnire, peng2025desire}, its conventional pipeline depends on expensive per-scene optimization, making it time-consuming and susceptible to shape--radiance ambiguities. To overcome these limitations and enable more efficient, scalable, and robust reconstruction, recent approaches \cite{jiang2025anysplat, miao2025evolsplat, yang2024storm, chen2025dggt} propose feedforward models that directly regress Gaussian parameters.

Despite their fast inference, these generalizable approaches suffer from fundamental limitations. First, the number of Gaussians is tightly coupled to the resolution of the input images. Capturing fine-grained details therefore requires a global increase in primitive count, leading to prohibitive memory consumption and effectively precluding the use of high-resolution training data. Notably, training DepthSplat \cite{xu2025depthsplat} with moderately high resolutions commonly used in practice (e.g., $1024 \times 1024$) already exceeds the memory capacity of high-end GPUs such as the NVIDIA A100. As a result, most existing methods are constrained to low input resolutions, typically $256 \times 256$. Second, consistently fusing Gaussians predicted from multiple views remains highly challenging. To mitigate cross-view inconsistencies and visual artifacts, current pipelines often limit the number of input views to a sparse subset. This issue is clearly observed in AnySplat \cite{jiang2025anysplat}, which fails to effectively exploit denser observations and exhibits degraded reconstruction quality when more than 32 input views are provided on its evaluation benchmarks.


To address these challenges, we draw inspiration from the core principles of 3DGS and recent advances in generalizable reconstruction \cite{chen2024g3r, wang2025flux4d, xu2025resplat}. We posit that generalizable reconstruction should not be treated as a one-shot regression problem, but instead formulated as an iterative refinement process. This perspective parallels the workflow of a sculptor, who does not obtain a final form by casting a single mold, but progressively shapes the artifact through repeated comparison between the evolving structure and its reference blueprint. Building on this insight, we introduce a robust framework for generalizable optimization and densification of 3D Gaussians.

However, extending this paradigm to dynamic urban scene reconstruction introduces several unique challenges. \textbf{(1) Ill-defined supervision for densification.} Determining where to densify Gaussians during training is inherently difficult. In iterative pipelines, the benefit of an individual densification operation is often delayed and implicit, only becoming apparent in the final reconstruction quality, which makes it challenging to design explicit and localized supervision signals. \textbf{(2) Irreversible degradation in early stages.} In complex street scenes, foreground Gaussians frequently elongate into needle-like structures to overfit high-frequency details, while low-frequency regions are erroneously compensated by the sky. Once such artifacts emerge, they are largely irreversible, severely limiting the effectiveness of subsequent refinement.

In this paper, we introduce \textbf{L2D2-GS}, a unified framework for generalizable reconstruction of dynamic Gaussian scenes that directly addresses these challenges. To resolve the ambiguity in densification supervision, we propose a policy-based learning strategy that derives explicit reward signals from the reconstruction process. Concretely, leveraging a differentiable renderer, we back-project global reconstruction improvements to individual densified Gaussians, enabling the densification policy to be trained in a fully self-supervised manner. To mitigate irreversible degradation in early optimization stages, we further propose a geometric regularization scheme tailored to generalizable optimization. By adopting a re-parameterization strategy that constrains the optimization manifold of Gaussian attributes, our method prevents primitives from collapsing into unfavorable local optima, thereby preserving structural fidelity during the highly unstable initial phases. The contributions of our work are summarized as follows.
\begin{itemize}[nosep]
    \item \textbf{Framework.} We introduce a learning-to-densify framework that enables generalizable reconstruction of dynamic urban scenes through a combination of feedforward optimization and adaptive densification.

\item \textbf{Method.} We propose a \textbf{self-supervised densification policy} that resolves supervision ambiguity by exploiting reconstruction gains to guide densification decisions. In addition, we introduce a \textbf{geometric regularization mechanism} based on reparameterization, which constrains the optimization manifold and effectively prevents early-stage geometric degradation.


\item \textbf{Evaluation.} Extensive experiments on common datasets demonstrate that our method significantly outperforms existing state-of-the-art approaches in both subjective visual quality and objective reconstruction metrics.

\end{itemize}

\section{Related Work}

\subsection{Feedforward 3D Gaussian Splatting}
Recent approaches bypass costly per-scene optimization by predicting Gaussian parameters in a feedforward manner. Early works \cite{charatan2024pixelsplat, chen2024mvsplat, zhang2025transplat, xu2025depthsplat} leverage multi-view stereo cues or transformer-based feature aggregation, while Large Reconstruction Models (LRMs) \cite{xu2024grm, zhang2024gs, ziwen2025long} scale this paradigm using large pre-trained networks to capture broader context. Extensions have further explored single-view reconstruction \cite{szymanowicz2024splatter}, panoramic settings \cite{chen2024mvsplat360}, pose-free formulations \cite{smart2024splatt3r, ye2024no}, and iterative refinement \cite{xu2025resplat}.

Most closely related to our work are methods targeting unconstrained or large-scale outdoor environments. In street scene reconstruction, EvolSplat \cite{miao2025evolsplat} and Omni-Scene \cite{wei2025omni} introduce volumetric or hybrid priors to maintain geometric consistency under sparse observations. Pushing further, AnySplat \cite{jiang2025anysplat} employs visually grounded geometric transformers for robust feedforward reconstruction, while STORM \cite{yang2024storm} explicitly models dynamic agents in large-scale scenes. However, these approaches either inadequately model dynamic objects or struggle to support high-resolution, long-duration reconstruction.

To bridge the gap between feedforward efficiency and the fidelity of per-scene optimization, recent works have explored \emph{generalizable optimization}. For instance, Splatformer \cite{chen2024splatformer} aggregates local geometric features via point transformers to improve out-of-distribution rendering quality. G3R \cite{chen2024g3r} proposes a gradient-guided paradigm that iteratively refines Gaussian parameters using rendering-loss gradients, enabling stronger generalization to unseen scenes. Extending this idea to the temporal domain, Flux4D \cite{wang2025flux4d} introduces a flow-based unsupervised framework that jointly optimizes geometry and motion, showing promising results for dynamic scene reconstruction.
Despite their strong generalization, these methods primarily focus on attribute refinement and therefore rely heavily on the quality of the initial point cloud \cite{chen2024g3r}. To overcome this limitation, we propose a reconstruction framework that jointly optimizes both attributes and geometry, enabling robust recovery from sparse or imperfect initialization.

\subsection{Adaptive Gaussian Densification}
Most existing Gaussian densification strategies are developed under per-scene optimization paradigms, where density is refined through gradient-based or photometric cues \cite{ye2024absgs, rota2024revising, kim2024color, Duan2026BUGS, Shuai2026Adversarial, Liang20264DGStream}. To cope with sparse observations and dynamic content, subsequent works introduce alternating optimization schemes \cite{patle2025ad}, explicit point cloud enhancement \cite{chan2024point}, or urban-scene-specific adaptations \cite{mohamad2025denser}. However, their reliance on iterative \emph{split-and-clone} operations makes them inherently incompatible with real-time or feedforward reconstruction pipelines.

More recently, learning-based approaches have explored single-pass densification. QuickSplat \cite{liu2025quicksplat} adopts lightweight networks for direct prediction, while Generative Densification \cite{nam2025generative} leverages large-scale priors to upsample Gaussian representations. GaussianLens \cite{weng2025gaussianlens} further proposes an on-demand strategy that augments coarse 3DGS with high-resolution local cues. Despite these advances, substantial limitations remain for large-scale reconstruction. Existing methods are largely designed for object-centric scenarios and exhibit severe memory bottlenecks when extended to long temporal sequences. As a result, they struggle to handle the complex topology and high-resolution requirements of unbounded driving environments. As summarized in Table~\ref{tab:comparison}, a clear gap remains for a generalizable densification framework capable of addressing these open-world challenges.

\subsection{3DGS for Autonomous Driving}
Applying 3D Gaussian Splatting to autonomous driving introduces distinct challenges stemming from large-scale scenes, dynamic agents, and ego-centric data sparsity. Early approaches primarily rely on per-scene optimization to construct high-fidelity digital twins. Street Gaussians \cite{yan2024street} and Omnire \cite{chen2024omnire} pioneer explicit decomposition of static backgrounds and dynamic objects, using 3D bounding boxes to separately track and reconstruct moving vehicles. To reduce reliance on manual annotations, Desire-GS \cite{peng2025desire} proposes a self-supervised framework that disentangles dynamic objects using appearance-based cues, while AD-GS \cite{xu2025ad} introduces object-aware B-spline Gaussians to model continuous trajectories. Addressing multi-sensor fusion, Splatad \cite{hess2025splatad} integrates LiDAR and camera data to account for real-world effects such as rolling shutter during reconstruction. Although these methods adapt Gaussian Splatting to the complexities of dynamic urban environments, they fundamentally inherit the iterative per-scene optimization paradigm of vanilla 3DGS, resulting in slow reconstruction speeds that limit their practicality in real-world autonomous driving systems.

\begin{table}[t]
    \centering
    \setlength{\tabcolsep}{4pt}
    
    \caption{\textbf{Feature comparison with SoTA methods.} We evaluate capabilities across Unbounded (Unb.) scenes, Dynamic (Dyn.) scenes, High-Resolution (High Res.) inputs, Density Control (Dens. Ctrl.), and Generalizability (Gen.).}
    \label{tab:comparison}
    
    \begin{tabular}{lccccc}
        \toprule
        Method & Unb. & Dyn. & \shortstack{High\\Res.} & \shortstack{Dens.\\Ctrl.} & Gen. \\
        \midrule
        \multicolumn{6}{l}{\textit{feedforward Densification}} \\
        Generative Densification \cite{nam2025generative} & & & & \cmark & \cmark \\
        GaussianLens \cite{weng2025gaussianlens} & & & \cmark & \cmark & \cmark \\
        QuickSplat \cite{liu2025quicksplat} & & & & \cmark & \cmark \\
        \midrule
        \multicolumn{6}{l}{\textit{Generalizable Prediction}} \\
        AnySplat \cite{jiang2025anysplat} & \cmark & & & & \cmark \\
        STORM \cite{yang2024storm} & \cmark & \cmark & & & \cmark \\
        \midrule
        \multicolumn{6}{l}{\textit{Generalizable Optimization}} \\
        G3R \cite{chen2024g3r} & \cmark & \cmark & \cmark & & \cmark \\
        Flux4D \cite{wang2025flux4d} & \cmark & \cmark & \cmark & & \cmark \\
        \midrule
        \multicolumn{6}{l}{\textit{Per-scene Optimization}} \\
        Omnire \cite{chen2024omnire} & \cmark & \cmark & \cmark & \cmark & \\
        StreetGS \cite{yan2024street} & \cmark & \cmark & \cmark & \cmark & \\
        \midrule
        \rowcolor{gray!15} \textbf{Ours} & \cmark & \cmark & \cmark & \cmark & \cmark \\
        \bottomrule
    \end{tabular}
\end{table}

\begin{figure*}[t]
    \centering
    \includegraphics[width=0.98\linewidth]{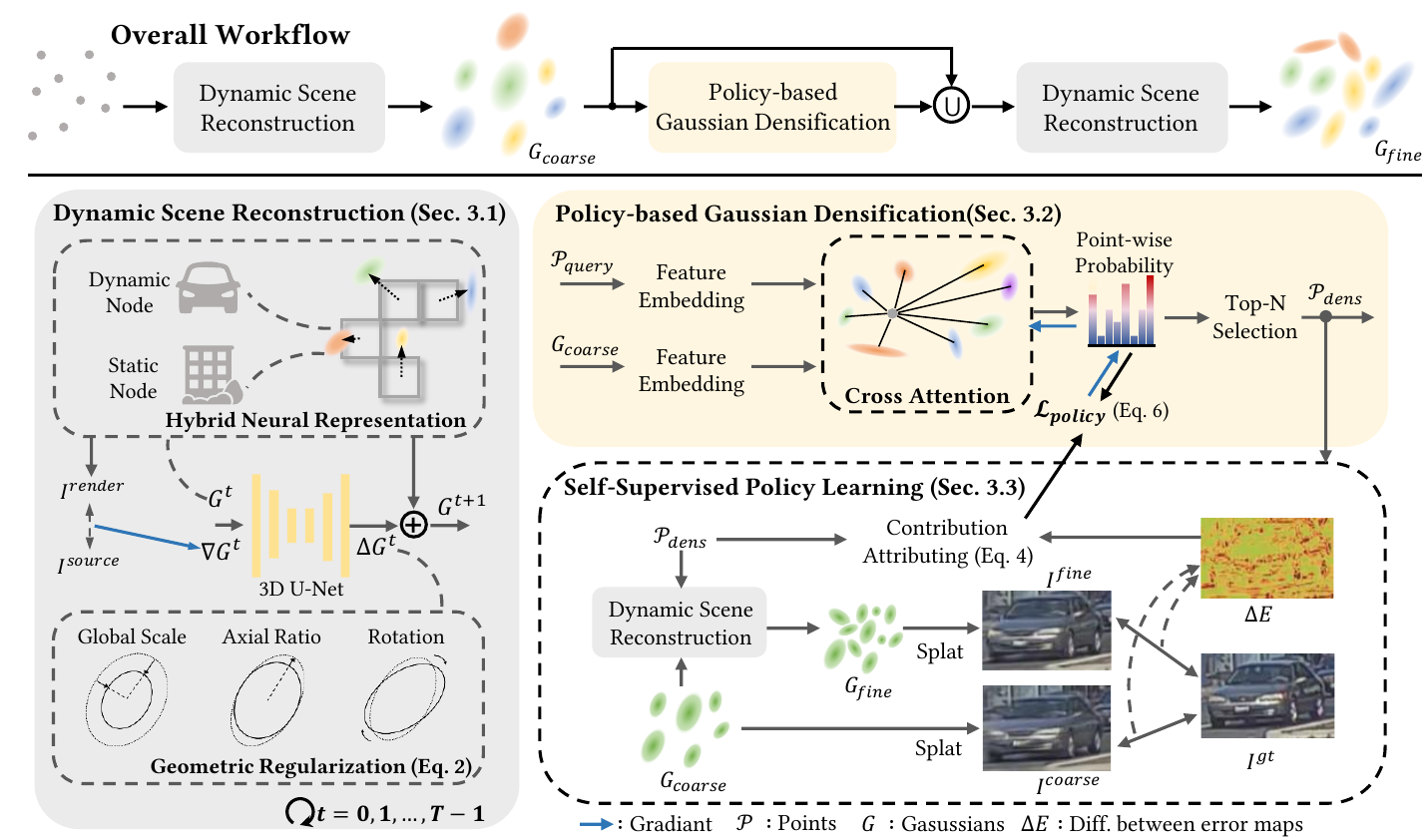}
    \caption{
    \textbf{Overview of the proposed framework.} Our method integrates a Dynamic Scene Reconstruction Module with a Policy-based Densification Module for dynamic outdoor scenes. 
       We employ a Hybrid Neural Representation that decouples the scene into static and dynamic nodes, modeled via voxel-wise neural Gaussians. These Gaussians are recurrently optimized by sparse 3D U-Net. 
      The Densification Module utilizes an attention mechanism to analyze local contexts of reconstructed Gaussians and predicts densification probabilities for query points. To supervise this process without ground truth, we propose a self-supervised learning strategy. This strategy computes an importance score for each densified point by weighting its pixel-wise visibility against the rendering error reduction.}
    \vspace{-10pt}
    \label{fig:pipeline}
\end{figure*}

\section{Methods}

Our proposed L2D2 framework operates by alternating between a feedforward reconstruction module and a policy-based densification module, as illustrated in Fig.~\ref{fig:pipeline}. To model complex dynamics, we adopt a scene-graph representation that decomposes urban scenes into modular static and dynamic nodes via predefined SE(3) transformations. Within the reconstruction module (Sec.~\ref{sec:scene_reconstruction}), each node is parameterized by voxel-wise neural Gaussians and recurrently optimized using 3D U-Nets equipped with sparse convolutions. Interleaved with this continuous refinement is the policy-based densification module (Sec.~\ref{sec:policy_densification}), which leverages an attention mechanism to identify geometrically under-represented regions and assigns spawning probabilities to candidate locations. To train this discrete point-selection process without ground-truth supervision, we introduce a self-supervised scoring mechanism (Sec.~\ref{sec:Self_Supervised}) that attributes the reduction in rendering loss to individually added Gaussians, providing a fine-grained reward signal. Finally, the complete training and inference pipeline orchestrating these components is detailed in Sec.~\ref{sec:training}.

\subsection{Dynamic Scene Reconstruction}
\label{sec:scene_reconstruction}

\noindent\textbf{Hybrid Neural Representation.}
We represent a dynamic scene as a composite scene graph composed of neural Gaussian primitives. Following G3R \cite{chen2024g3r}, each Gaussian is parameterized as $\mathcal{G} = \{\Theta, \mathbf{f}\}$, where $\Theta$ denotes explicit attributes—including position $\mu$, rotation $q$, scale $s$, color $c$, and opacity $\alpha$—and $\mathbf{f}$ is a latent neural feature. A shallow MLP decoder $\Phi$ maps these parameters to rectified attributes used for differentiable rendering. For efficient network processing, the Gaussians are initially organized into a sparse voxel grid, with exactly one primitive assigned to each voxel.

To model the complex dynamics of urban scenes, we instantiate three types of scene-graph nodes:  
(i) \textit{Static background nodes}, defined in the world coordinate frame;  
(ii) \textit{Dynamic rigid nodes}, defined in a canonical object space and transformed over time using ground-truth trajectories $\mathbf{T}_t = [\mathbf{R}_t \mid \mathbf{t}_t]$; and  
(iii) a learnable \textit{sky cubemap} that represents distant background content.  
The final image is synthesized through unified differentiable Gaussian splatting, compositing foreground Gaussians over the sky background.

\vspace{1pt}
\noindent\textbf{Iterative Feedforward Optimization.}
We formulate scene reconstruction as a learnable iterative feed-forward process parameterized by a Sparse 3D U-Net $\Psi$. At each iteration $t$, the current scene representation is rendered from the input viewpoints, and gradients of the Gaussian parameters with respect to the photometric MSE loss are computed. These gradients serve as error-feedback signals and are channel-wise concatenated with the current Gaussian attributes $\mathcal{G}_t$ before being fed into $\Psi$. The network predicts a residual update that refines the scene representation:
\begin{equation}
    \mathcal{G}_{t+1} = \mathcal{G}_t + \Psi\!\left(\mathcal{G}_t, \nabla \mathcal{G}_t \right).
\end{equation}

In parallel, a lightweight two-layer residual convolutional network is used to iteratively reconstruct the sky cubemap. The entire framework is trained end-to-end by minimizing the rendering loss of the scene at each step.


\vspace{1mm}
\noindent\textbf{Geometric Regularization via Reparameterization.}
We observe that naively applying the proposed framework leads to severe geometric degeneration, particularly during the early iterations of optimization. In practice, a pathological failure mode arises in which foreground Gaussians rapidly collapse into extreme, needle-like shapes to overfit high-frequency image residuals, while low-frequency components are erroneously compensated by background Gaussians. Once the optimization enters this regime, a vicious cycle ensues: the distorted geometry produces misleading rendering gradients that no longer provide meaningful guidance for shape recovery, rendering subsequent refinement ineffective. To mitigate this issue, we introduce a geometric re-parameterization that regularizes the optimization process.

Instead of treating the three scaling axes as independent variables, we decouple them into a single \textit{global log-scale} $s_{g} \in \mathbb{R}$ and two \textit{expansive ratio} coefficients $\rho = [\rho_1, \rho_2] \in \mathbb{R}^2$. The final scaling vector $\mathbf{s}$ is derived via specific activation functions:

\begin{equation}
    \mathbf{s} = \phi(s_{g}) \cdot [1, \psi(\rho_1), \psi(\rho_1)\psi(\rho_2)]^T,
\end{equation}

here $\phi(x) = \exp(x)$ controls the global scale magnitude, while $\psi(x) = |x| + 1$ parameterizes the degree of shape anisotropy. This design constrains the solution space to a manifold with an isotropic boundary. Initializing $\rho = 0$ anchors each Gaussian at the isotropic limit of the manifold, since $\psi(0) = 1$. As a result, the network is prevented from arbitrarily stretching Gaussian primitives; instead, it must learn a well-defined trajectory on the constrained manifold to introduce anisotropy. This structural constraint acts as a strong regularizer, effectively preventing needle-like degeneracies during the early stages of optimization. A qualitative visualization illustrating the effect of this design is provided in the supplementary material.


\subsection{Policy-based Gaussian Densification}
\label{sec:policy_densification}
In large-scale outdoor environments, dense geometric priors are typically obtained either from dedicated hardware sensors, such as LiDAR, or from large vision foundation models (e.g., VGGT \cite{wang2025vggt}). However, naively initializing 3D Gaussian Splatting with such massive point clouds---where concatenating LiDAR frames of a medium-sized scene can easily exceed $10^7$ points---incurs prohibitive memory costs that surpass the capacity of even high-end GPUs.

To balance reconstruction fidelity with computational efficiency, we propose a densification strategy driven by a generalizable network. Rather than relying on brute-force initialization, our approach follows a \emph{grow-on-demand} paradigm: reconstruction begins from a sparse geometric initialization, and the network progressively identifies and populates under-represented regions with new Gaussian primitives during the iterative optimization process.
Since the insertion of Gaussian primitives is a discrete and inherently non-differentiable operation, we formulate densification as a learnable policy. Inspired by reinforcement learning, we treat the coarse Gaussian reconstruction $\mathcal{G}_{\text{coarse}}$ as the current \textbf{state}, while the densification \textbf{action} is represented by a continuous probability distribution $\{w_i\}$, where each $w_i$ reflects the policy’s confidence in incorporating a candidate primitive into the scene.

To instantiate the action space, we construct a candidate set $\mathcal{P}_{\text{query}}$ by integrating multi-modal geometric priors (e.g., LiDAR points and depth-based unprojection). The policy network embeds both the candidate primitives and $\mathcal{G}_{\text{coarse}}$ into separate feature spaces and applies a $k$-Nearest Neighbor ($k$-NN) attention mechanism to capture local geometric context. By aggregating structural information from spatial neighbors, the network predicts a probability $w_i \in [0,1]$ for each candidate.
During the forward pass, we execute the policy using deterministic top-$N$ selection, instantiating only the highest-scoring candidates to efficiently alleviate geometric sparsity. To enable end-to-end learning despite the discrete decision process, we further derive a pseudo-ground-truth reward through point-wise scoring, as detailed in Sec.~\ref{sec:Self_Supervised}.

\subsection{Self-Supervised Policy Learning}
\label{sec:Self_Supervised}

Here, we detail the learning paradigm of the proposed densification policy.
To quantitatively evaluate the effectiveness of densification actions, we introduce a self-supervised training scheme that leverages feedback from the reconstruction module (Sec.~\ref{sec:scene_reconstruction}) to update the policy.
Specifically, we perform a tentative reconstruction using the densified scene state and measure the resulting improvement in rendering quality. The reduction in rendering loss is then attributed back to the newly added Gaussian primitives, producing a point-wise contribution score $S_i$ for each candidate. This score serves as a proxy supervision signal that reflects the candidate's contribution to the final image fidelity.
Finally, we formulate the policy loss by integrating these contribution scores with a geometric consistency regularization term. This joint objective trains the policy network to prioritize candidates that yield both photometric improvements and structurally plausible geometry, resulting in stable and effective densification.

\vspace{1mm}
\noindent\textbf{Attribution of Rendering Gains.}
We quantify the geometric utility of added Gaussians by measuring the reduction in rendering error between the coarse ($E_c$) and densified ($E_f$) reconstruction. Specifically, we calculate the per-pixel $L_2$ error maps for both passes, yielding a gain map $\Delta E = E_c - E_f$ that highlights regions with fidelity improvements.

To attribute these 2D pixel-wise gains back to specific 3D primitives, we leverage the differentiable rasterization pipeline. Specifically, we augment the densified Gaussians with an auxiliary mask channel $\mathbf{m}$ ($m_i=1$ for candidate points) and rasterize it to obtain a contribution map $M_{dens}^v$. We then treat the gain map $\Delta E$ as an upstream ``pseudo-gradient" signal and back-propagate it through a surrogate attribution loss:
\begin{equation}
    \mathcal{L}_{attr}^v = \langle \Delta E^v, M_{dens}^v \rangle.
\end{equation}
Intuitively, this inner product projects the 2D fidelity improvements onto the visibility footprint of the added geometry. By performing a standard backward pass, we derive the {rendering contribution score} $S_i^v$ as the gradient with respect to the mask input:
\begin{equation}
\label{eq:score}
    S_i^v \triangleq \nabla_{m_i} \mathcal{L}^v_{attr}.
\end{equation}
This metric serves as a sensitivity proxy, quantifying the individual contribution of each primitive to the global error reduction. Crucially, this process incurs zero additional computational overhead by directly reusing the optimized CUDA backward kernels.

\vspace{1mm}
\noindent\textbf{Policy Learning via Score Maximization.}
Drawing inspiration from reinforcement learning, we formulate the densification task as a utility maximization problem. For each candidate point, we compute a view-dependent utility score $u_i^v$ that balances its rendering contribution $S_i^v$ against a geometric consistency penalty:
\begin{equation}
    u_i^v = S_i^v - \lambda_{depth} \left| D_{i}^v - \hat{D}_{i}^v \right|.
\end{equation}
Here, $D_{i}^v$ represents the rendered surface depth at the projection of Gaussian $i$, while $\hat{D}_{i}^v$ denotes the center depth of the primitive itself. This term penalizes primitives that drift away from the reconstructed surface.

The objective is to maximize the expected utility $\mathcal{A}$ while enforcing the diversity of the decision policy via entropy maximization. We minimize the negative sum to perform this maximization:

\begin{equation}
\label{eq:policy_loss}
    \mathcal{L}_{policy} = - \frac{1}{K} \sum_{i=1}^K \left( \underbrace{w_i \sum_{v} u_i^v}_{\text{Expected Utility } \mathcal{A}} + \lambda_{e} \underbrace{\mathcal{H}(w_i)}_{\text{Entropy Loss } \mathcal{L}_{e}} \right),
\end{equation}
where $w_i$ is the selection probability predicted by the policy. Here, $\mathcal{H}(w_i)$ is the standard binary entropy defined as:
\begin{equation}
    \mathcal{H}(w_i) = - \left[ w_i \log w_i + (1-w_i) \log (1-w_i) \right].
\end{equation}
By maximizing this entropy term, we penalize over-confident predictions. This prevents the policy from converging to a deterministic state, thereby facilitating broader exploration of the geometric configuration space and stablizeing the training.

\subsection{Training \& Inference Process}
\label{sec:training}
The complete scene reconstruction process is outlined in Algorithm \ref{alg:training_pipeline}, which alternates between continuous Gaussian parameter updates via the reconstructor and discrete point expansion driven by the policy network. However, simultaneously optimizing both components from a cold start presents significant convergence challenges. Therefore, to ensure stability and efficiency, our training pipeline is divided into three distinct phases: (1) Reconstructor Pre-training, (2) Policy Learning, and (3) Joint Training.

\noindent\textbf{Stage 1: Reconstructor Pre-training.}
We first optimize the reconstruction network 
(Sec. \ref{sec:scene_reconstruction}).
Same to Lines 5--10 of Algorithm \ref{alg:training_pipeline}, this process involves an iterative update mechanism: the network refines the Gaussian attributes based on their current state and the gradients back-propagated from the differentiable rasterizer. To train the network, we employ a combination of L2 photometric loss, LPIPS perceptual loss, and a scale regularization term to the optimized Gaussians:
\begin{equation}
    \mathcal{L}_{\text{recon}} = \| I_{\text{ren}} - I_{\text{gt}} \|_2 + \lambda_{\text{lpips}} \mathcal{L}_{\text{LPIPS}}(I_{\text{ren}}, I_{\text{gt}}) + \lambda_{\text{reg}} \mathcal{L}_{\text{reg}}(\mathcal{G}),
\end{equation}
where $I_{\text{ren}}$ and $I_{\text{gt}}$ denote the rendered and ground-truth images, respectively. The regularization term $\mathcal{L}_{\text{reg}}$ is explicitly designed to penalize excessively large Gaussians, preventing artifacts caused by oversized primitives:
\begin{equation}
    \mathcal{L}_{\text{reg}}(\mathcal{G}) = \sum \max(0, \phi(s_{g}) - \epsilon),
\end{equation}
where $\phi(s_{g})$ represents the global scale for the $i$-th Gaussian, and $\epsilon$ serves as a scale threshold.

    
    

\begin{algorithm}[t]
\caption{L2D2-GS Pipeline}
\label{alg:training_pipeline}
\begin{algorithmic}[1]
\Require Sparse points $\mathcal{P}_{\text{sparse}}$, Candidate set $\mathcal{P}_{\text{query}}$
\Require Input Images $\mathbf{I}$, Camera Poses $\Pi$
\Require Policy network $\pi$, Reconstruction network $\Psi$

\State \textbf{Initialization:}
\State $\mathcal{P}_{\text{curr}} \gets \mathcal{P}_{\text{sparse}}$
\State Initialize Gaussians $\mathcal{G}^{(0)}$ from $\mathcal{P}_{\text{curr}}$ \Comment{Zero-init features}

\State \Comment{\textbf{Coarse Reconstruction}}
\For{$t = 0, 1, \dots, T_1 - 1$}
    \State $\hat{\mathbf{I}} \gets f_{\text{rast}}(\mathcal{G}^{(t)}, \Pi)$ \Comment{Render images}
    \State $\nabla_{\mathcal{G}^{(t)}} \gets \nabla \| \hat{\mathbf{I}} - \mathbf{I} \|_2$ \Comment{Compute gradients via rendering loss}
    \State $\mathcal{G}^{(t+1)} \gets \mathcal{G}^{(t)}+\Psi_t(\mathcal{G}^{(t)}, \nabla_{\mathcal{G}^{(t)}})$ \Comment{Forward update pass}
    \If{training}
        \State Update $\Psi_t$ via $\mathcal{L}_{\text{recon}}$
    \EndIf
\EndFor

\State \Comment{\textbf{Densification Step}}
\State $w \gets \pi(\mathcal{P}_{\text{query}}, \mathcal{G}^{(T_1)})$ \Comment{Generate probabilities}
\State $\mathcal{P}_{\text{dens}} \gets \text{Select}(\mathcal{P}_{\text{query}}, w)$ \Comment{Select points based on current policy}
\State $\mathcal{P}_{\text{curr}} \gets \mathcal{P}_{\text{curr}} \cup \mathcal{P}_{\text{dens}}$ \Comment{Augment geometry}
\State Initialize Gaussians $\mathcal{G}^{(0)}$ from $\mathcal{P}_{\text{curr}}$ 

\State \Comment{\textbf{Fine Reconstruction \& Policy Update}}
\For{$t = 0, 1, \dots, T_2 - 1$}
    \State $\hat{\mathbf{I}} \gets f_{\text{rast}}(\mathcal{G}^{(t)}, \Pi)$
    \State $\nabla_{\mathcal{G}^{(t)}} \gets \nabla \| \hat{\mathbf{I}} - \mathbf{I} \|_2$
    \State $\mathcal{G}^{(t+1)} \gets \mathcal{G}^{(t)}+\Psi_t(\mathcal{G}^{(t)}, \nabla_{\mathcal{G}^{(t)}})$ 
    
    \If{training}
        \State Update $\Psi_t$ via $\mathcal{L}_{\text{recon}}$
        
        \If{$t == T_1 - 1$} \Comment{Policy Update}
            \State Compute scores $S$ for $\mathcal{P}_{\text{dens}}$
            \State Update Policy $\pi$ via $\mathcal{L}_{\text{policy}}(S, w)$
        \EndIf
    \EndIf
\EndFor

\end{algorithmic}
\end{algorithm}

\vspace{1mm}
\noindent\textbf{Stage 2: Policy Learning from Cached Scores.}
Calculating the gradient-based score $S$ (Eq. \ref{eq:score}) requires two reconstruction passes of the scene, which is computationally expensive to perform at every iteration. To accelerate training, we implement an offline caching strategy in this stage. Specifically, we pre-compute and store data tuples in the form of $\{ \mathcal{G}_{\text{coarse}}, \mathcal{P}_{\text{dens}}, S \}$, comprising the current coarse Gaussians, the candidate points, and their evaluated scores. Thereby, the policy network is efficiently trained via the policy loss (Eq. \ref{eq:policy_loss}) using these pre-calculated supervision signals.

\vspace{1mm}
\noindent\textbf{Stage 3: Joint Training.}
Finally, we unfreeze all modules and jointly optimize the entire framework according to Algorithm \ref{alg:training_pipeline}. This stage allows the reconstruction network to adapt to the newly densified geometry, while the policy can update based on the improved rendering feedback. The optimization objective is consistent with the previous phases.


\begin{figure*}
    \centering
    \includegraphics[width=\linewidth]{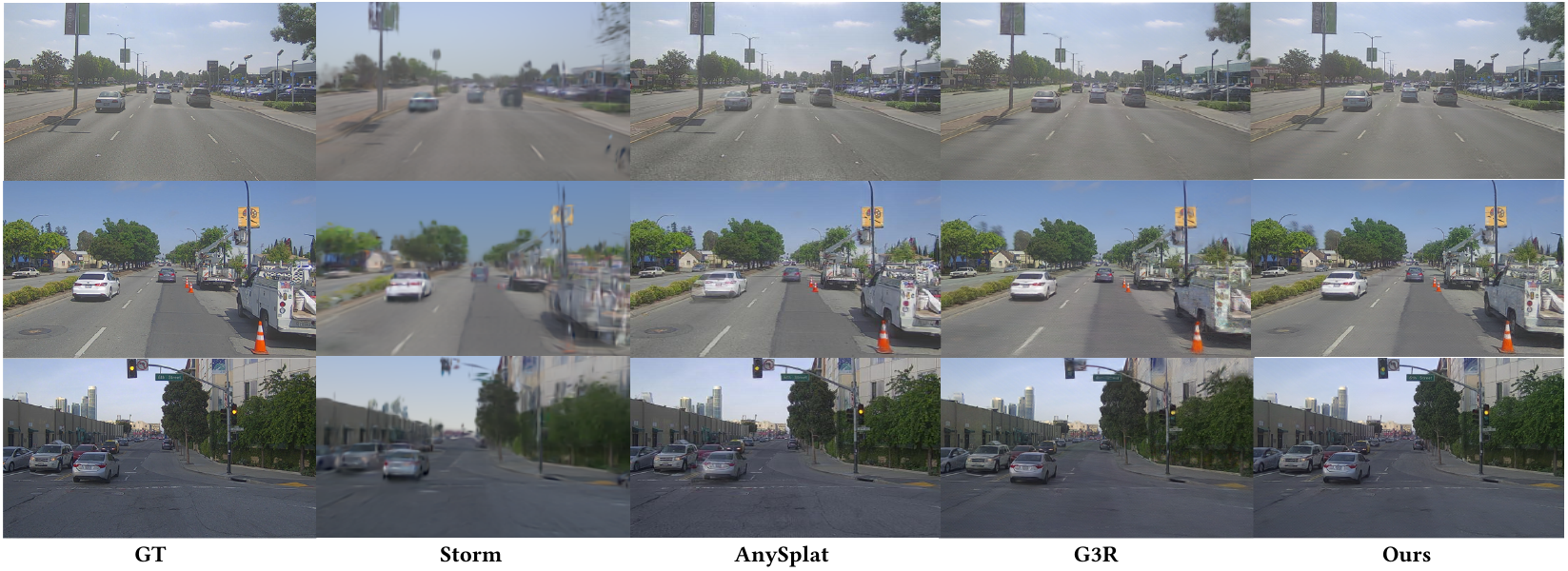}
    \vspace{-15pt}
    \caption{\textbf{Visual comparison on Pandaset.} Our approach produces sharp results with less artifacts, whereas competitors struggle to handle dynamics, exhibiting issues like distorted foreground objects or significant blurring.}
    \label{fig:pandaset_1s}
\end{figure*}

\begin{figure*}
    \centering
    \includegraphics[width=\linewidth]{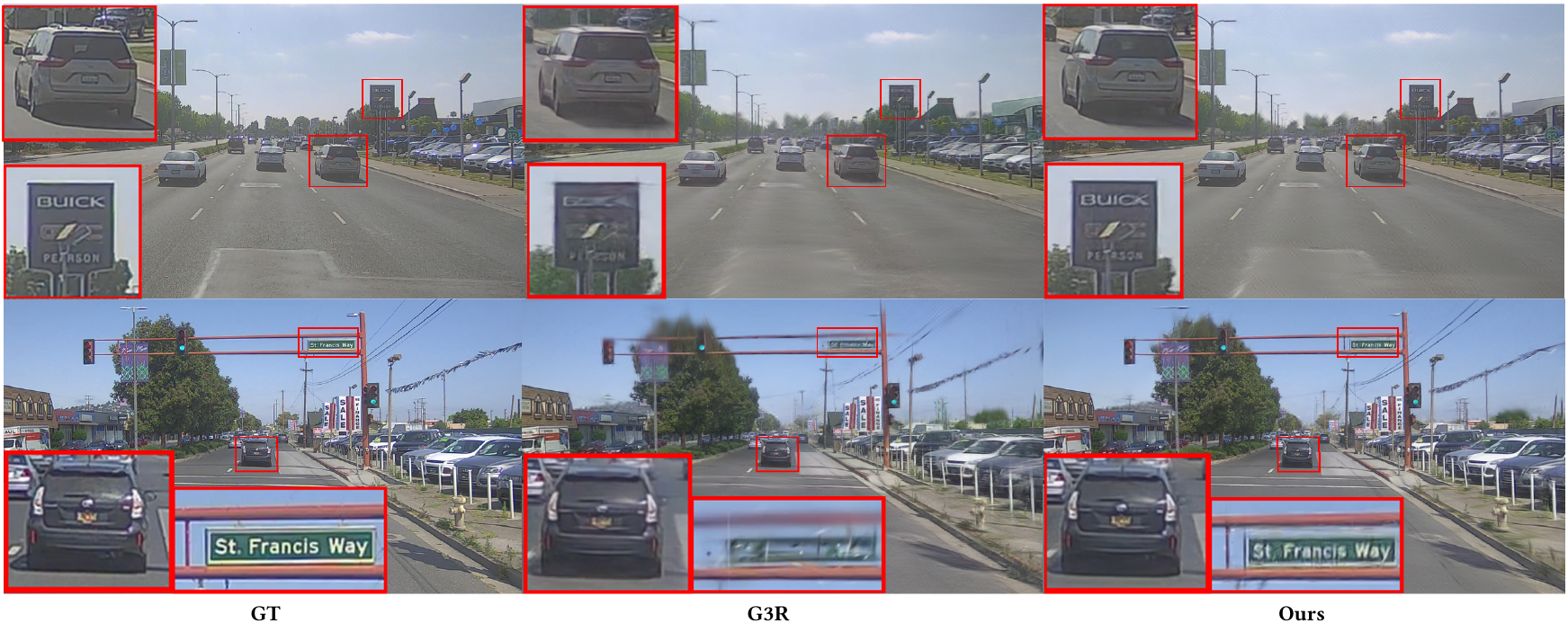}
    \vspace{-15pt}
    \caption{\textbf{Full sequence reconstruction on Pandaset.} While G3R exhibits severe blurring and artifacts due to missing geometry, our method preserves sharp textures and clear geometry, validating the efficiency of our self-supervised densification strategy.}
    \vspace{-10pt}
    \label{fig:pandaset_8s}
\end{figure*}

\section{Experiments}

\subsection{Implementation Details}
\label{sec:implementation}

\noindent\textbf{Network Architecture.}
Our reconstructor builds upon the encoder-decoder SparseResUNet architecture from \texttt{torchsparse} \cite{tang2023torchsparse++}. The network takes the concatenation of 3D Gaussian features and rendering-induced gradients as its input. To effectively accommodate distribution shifts across different optimization iterations, we employ step-specific independent weights rather than a shared network. The final residual updates are then predicted via a lightweight output head consisting of a single linear layer with a \texttt{tanh} activation. Furthermore, to enable a coarse-to-fine reconstruction paradigm, we apply a linear scheduling strategy that progressively decays the update rate over iterations, thereby stabilizing the optimization and facilitating the refinement of high-frequency details. Finally, the sky cubemap ($512 \times 512$ resolution per face) modeling distant backgrounds is explicitly optimized using a two-layer residual convolutional network.


The policy network is parameterized by a $k$-NN cross-attention architecture. To handle heterogeneous inputs, independent two-layer MLPs first embed candidate points and existing Gaussians into $64$-dimensional latent spaces. A $k$-NN cross-attention module ($k=16$, $N_{\text{heads}}=4$, $d_{\text{head}}=32$) then aggregates context by allowing candidate points to attend to their local Gaussian neighbors. The retrieved context features are concatenated with the original candidate embeddings and fed into a scoring MLP, followed by a Sigmoid activation to predict the discrete densification probability $w \in [0,1]$.

\noindent\textbf{Inference Configuration.}
We initialize with a set of $|\mathcal{P}_{\text{sparse}}| = 800\text{K}$ points.
For the reconstruction process, we set the number of coarse and fine optimization steps to $T_1=12$ and $T_2=24$, respectively. 
For the densification phase, the policy selects top-$N$ points from a candidate pool of $|\mathcal{P}_{\text{query}}| = 4\text{M}$ points to reach $|\mathcal{P}_{\text{dens}}| = 400\text{K}$. The candidates are sampled from the dataset's raw LiDAR scans and geometric priors derived from MapAnything~\cite{keetha2025mapanything}. 

\noindent\textbf{Training Configuration.}
We implement our framework using PyTorch. All models are trained with 8 $\times$ NVIDIA H20 GPUs. We set the per-GPU scene batch size to 1, resulting in an total scene batch size of 8.
We utilize the {AdamW} optimizer with an initial learning rate of $1 \times 10^{-4}$ and a weight decay of $1 \times 10^{-5}$. The loss weights are empirically set as follows: the perceptual loss weight $\lambda_{lpips} = 0.01$, the regularization weight $\lambda_{reg} = 0.01$, and the scale threshold $\epsilon = 0.01$. For the policy-based densification module, we set the depth penalty weight $\lambda_{depth} = 0.01$ and the entropy constraint weight $\lambda_{e} = 0.02$.

\noindent\textbf{Datasets.}
We conduct our primary evaluation on the {PandaSet} dataset \cite{xiao2021pandaset}, which features challenging dynamic urban scenes captured with high-resolution LiDAR and cameras. To ensure a fair and consistent comparison with SoTA methods, we strictly adhere to the data partition protocol used in Flux4D \cite{wang2025flux4d}.
To further assess the robustness of our reconstruction framework, we perform a cross-dataset generalization test on the {Waymo Open Dataset} (WOD) \cite{sun2020scalability}. In this setting, we directly evaluate the model trained on PandaSet on unseen Waymo sequences without any fine-tuning. We use the NOTR Dynamic32 split provided by Emernerf \cite{yang2023emernerf} for evaluation.

\subsection{Main Results}

\noindent\textbf{Full Sequence Reconstruction.}
Table \ref{tab:full_seq_interpolation} reports the performance on full sequence reconstruction (80 frames), where odd frames are used as inputs and even frames serve as the target novel views for evaluation. 
We include per-scene optimization methods \cite{wei2025omni, yan2024street} as an upper-bound reference. While they yield higher PSNR, their deployment suffers from prohibitive reconstruction time ($\sim$70 mins per scene).

In the realm of generalizable methods, our approach achieves substantial rendering gains and superior visual fidelity. Compared to G3R \cite{chen2024g3r}, we deliver higher rendering quality while maintaining a compact representation. Qualitatively, as shown in Fig.~\ref{fig:pandaset_8s}, this rendering advantage is clearly pronounced. While G3R suffers from noticeable blurring and visual artifacts, our method reconstructs sharper novel views and faithfully preserves fine-grained details. This photometric improvement stems from our policy-based densification strategy, which explicitly allocates primitives to maximize rendering rewards. Furthermore, to push the limits of efficiency, we evaluate a fast variant (\textbf{Ours-f}) configured with reduced optimization steps and smaller candidate pool ($T_1=4$, $T_2=8$, $|\mathcal{P}_{\text{query}}|=2\text{M}$). This setup significantly reduces the computational overhead with marginal performance degradation. Discussion regarding optimization steps is provided in the supplementary material.


\begin{table}[t]
    \centering
    \caption{\textbf{Quantitative comparison on full sequence interpolation.} \textbf{Bold} and \underline{underline} indicate the best and second-best results among generalizable approaches, respectively. '*' indicates reproduced results.}
    \label{tab:full_seq_interpolation}
    {
    \begin{tabular}{l|ccc|cc}
    \toprule
    Method & PSNR $\uparrow$ & SSIM $\uparrow$ & LPIPS $\downarrow$ & Time & $N_{gs}$ \\
    \midrule
    \multicolumn{6}{l}{\textit{\textbf{Optimization}}} \\
    StreetGS & 24.54 & 0.739 & 0.224 & $\sim$70min & 3M \\
    OmniRe   & 24.57 & 0.739 & 0.222 & $\sim$80min & 3M \\
    \midrule
    \multicolumn{6}{l}{\textit{\textbf{Generalizable}}} \\
    G3R      & 23.15 & 0.636 & -     & \underline{60s} & 3M \\
    G3R$^*$  & 23.18 & 0.653 & 0.406 & 75s & 3M \\
    \textbf{Ours}   & \textbf{24.19} & \textbf{0.705} & \textbf{0.329} & 98s & \textbf{1.2M} \\
    \textbf{Ours-f} & \underline{23.93} & \underline{0.703} & \underline{0.356} & \textbf{39s} & \textbf{1.2M} \\
    \bottomrule
    \end{tabular}
    }
\end{table}


\noindent\textbf{Short Sequence Reconstruction.}
We evaluate our method on short sequence interpolation following the protocol of Flux4D \cite{wang2025flux4d}, testing with 1s snippets. As detailed in Table \ref{tab:short_seq}, our approach consistently outperforms existing SoTA methods. Notably, we achieve a PSNR of 25.22 dB, surpassing the closest competitor G3R. However, we observe that the improvement in performance is smaller than in the full-sequence scenario, as inaccurate geometry in short snippets is easier to compensate for via overfitting.
Qualitatively, the superiority of our method remains evident. As illustrated in Fig.~\ref{fig:pandaset_1s}, our approach produces visual results with fewer artifacts compared to G3R. In contrast, other generalizable baselines struggle to handle scene dynamics: AnySplat fails to correctly model moving objects, leading to missing or distorted actors, while STORM suffers from severe motion blur.


\noindent\textbf{Future Extrapolation.}
We evaluate its performance on extrapolating to future frames beyond the training horizon. As illustrated in Fig. 8, our method generates plausible and temporally consistent views for these unseen frames. This demonstrates that our model has learned a generalized scene representation rather than merely overfitting to input views.

\begin{figure*}
    \centering
    \includegraphics[width=1\linewidth]{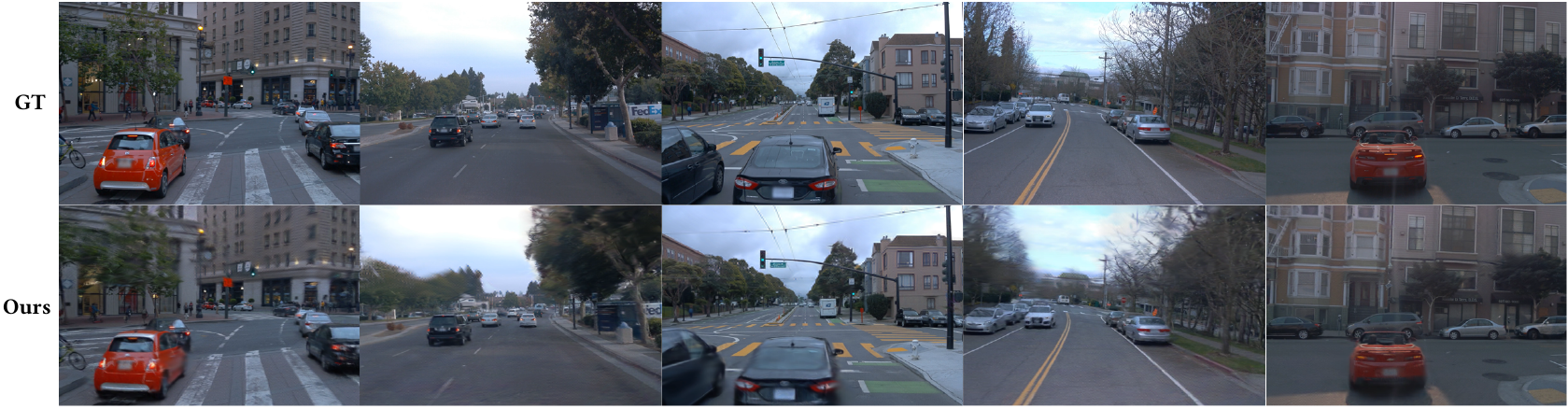}
    \caption{\textbf{Zero-shot generalization on Waymo Open Dataset.} We directly apply the model trained on PandaSet to Waymo sequences. Despite the domain gap, our method synthesizes high-fidelity novel views with consistent geometry, demonstrating robust cross-dataset generalization capabilities.}
    \vspace{-10pt}
    \label{fig:cross_dataset}
\end{figure*}

\noindent\textbf{Cross-Dataset Generalization.}
We evaluate generalizability of our framework by training on PandaSet and directly testing on the WOD using 1s-snippets. As detailed in Table \ref{tab:cross_dataset}, our method outperforms the baseline STORM, achieving $\sim$5 dB gain in PSNR and 0.19 reduction in LPIPS. As visualized in Fig.~\ref{fig:cross_dataset}, our method successfully reconstructs coherent scene structures even under domain shifts.

\begin{table}[t]
    \centering
    \caption{Quantitative comparison on short sequence interpolation. Since most feedforward methods cannot handle long dynamic sequences, Best results are \textbf{bolded}.}
    \label{tab:short_seq}
    \begin{tabular}{l c c c}
        \toprule
        Method & PSNR $\uparrow$ & SSIM $\uparrow$ & LPIPS $\downarrow$ \\
        \midrule
        AnySplat & 22.97 & 0.673 & 0.412 \\
        STORM & 19.69 & 0.628 & 0.683 \\
        G3R & 24.35 & 0.686 & - \\
        G3R* & 24.36 & 0.698 & 0.347\\
        Flux4D & 23.84 & 0.675 & - \\
        \midrule
        \textbf{Ours} & \textbf{25.22} & \textbf{0.735} & \textbf{0.287} \\
        \bottomrule
    \end{tabular}
\end{table}

\begin{table}[t]
    \centering
    \caption{\textbf{Quantitative comparison on cross-dataset generalization.} Our method demonstrates superior generalizability.}
    \label{tab:cross_dataset}
    \begin{tabular}{l|ccc}
        \toprule
        Method & PSNR $\uparrow$ & SSIM $\uparrow$ & LPIPS $\downarrow$ \\
        \midrule
        STORM & 20.14 & 0.734 & 0.630 \\
        \textbf{Ours} & \textbf{25.24} & \textbf{0.794} & \textbf{0.436} \\
        \bottomrule
    \end{tabular}
    
\end{table}

\subsection{Ablation Study}

\noindent\textbf{Component-wise Ablation.}
To validate the effectiveness of our design, we perform a component-wise ablation study as summarized in Table \ref{tab:ablation}. Specifically, for the variant w/o Densification, we increased the initialization density to ensure a final budget of 1.2 million Gaussians, matching the full model. Despite the equivalent primitive count, the rendering quality still drops (e.g., PSNR 24.19 $\to$ 23.78 dB), demonstrating that our adaptive densification strategy is far more effective than dense initialization in capturing high-frequency details. The performance degrades further without the reparameterization module (23.41 dB), confirming that both our proposed modules are essential for high-fidelity reconstruction. Finally, both replacing the step-wise independent weights with shared weights and removing the consistency penalty causes performance drop.


\begin{figure*}[t]
    \centering
    \includegraphics[width=\linewidth]{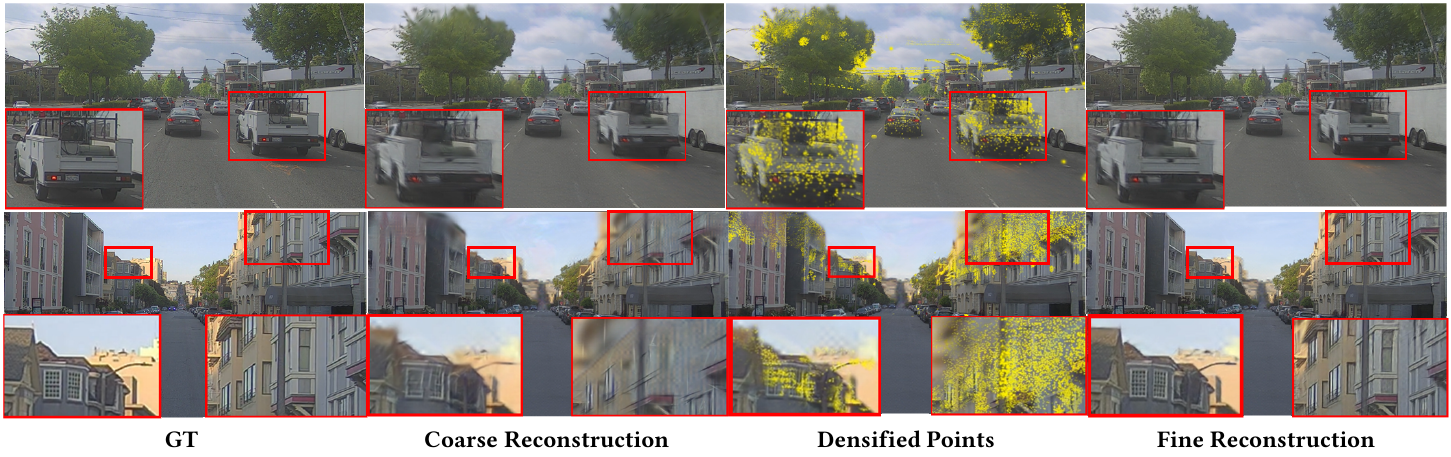}
    \caption{\textbf{Visualization of the densification process.} Our policy-based strategy effectively localizes under-represented areas within the scene that lack geometric detail. The newly added Gaussian primitives, highlighted as yellow points, are strategically inserted into these regions to refine the reconstruction.}
    \vspace{-10pt}
    \label{fig:densification}
\end{figure*}

\begin{table}[t]
    \centering
    \caption{\textbf{Component-wise ablation study.} We evaluate the impact of different core components of our proposed framework. To obtain reliable results, We evaluate the mean value ($\mu$) and standard deviation ($\sigma$) of all metrics by independently reconstructing each test scene $10$ times.}
    \label{tab:ablation}
    \begin{tabular}{l | c | c | c}
    \toprule
    \multirow{2}{*}{Method} & PSNR $\uparrow$ & SSIM $\uparrow$ & LPIPS $\downarrow$ \\
    & $\mu$ ($\sigma$ {\scriptsize$\times 10^{3}$}) & $\mu$ ($\sigma$ {\scriptsize$\times 10^{5}$}) & $\mu$ ($\sigma$ {\scriptsize$\times 10^{5}$}) \\
    \midrule
    \textbf{L2D2-GS} & \textbf{24.19 (1.96)} & \textbf{0.705 (8.62)} & \textbf{0.329 (6.46)} \\
    $-$ consistency penalty & 24.17 (1.78) & 0.704 (7.80) & 0.330 (7.72) \\
    $-$ densification       & 23.78 (1.86) & 0.682 (7.31) & 0.368 (6.54) \\
    $-$ reparameterization  & 23.41 (1.62) & 0.667 (6.87) & 0.396 (7.53) \\
    $-$ independent weights  & 23.18 (1.75) & 0.653 (8.21) & 0.406 (6.82) \\
    \bottomrule
    \end{tabular}
    
\end{table}


\noindent\textbf{Impact of Geometric Hypotheses.}
We investigate the influence of different point sources on our densification strategy, which is designed to be compatible with various geometric priors. As shown in Table \ref{tab:ablation_candidates}, relying solely on LiDAR provides high geometric accuracy but often suffers from sparsity, while using 3D-VFM improves scene coverage but may lack precision. By integrating both, our method effectively combines the geometric accuracy of LiDAR with the extensive scene coverage of 3D-VFM. This hybrid approach mitigates the limitations of individual sources, resulting in the best reconstruction quality across all metrics.

\noindent\textbf{Visualization of the Densification process.} To intuitively validate the efficacy of our strategy, we visualize the densified points in Fig. \ref{fig:densification}. As observed, the learned policy demonstrates a strong capability to localize under-represented regions. The densified Gaussians precisely align with geometric voids and structures that are insufficiently reconstructed. By strictly confining the growth of new primitives to these identified geometric gaps, our method effectively enhances reconstruction completeness while avoiding redundant Gaussians.

\noindent\textbf{Effect of Reparameterization.}
As illustrated in Fig.~7, our reparameterization approach substantially improves reconstruction quality by effectively reducing geometric degradation in the foreground. While the baseline suffers from needle-like artifacts, our method maintains clean and coherent geometry in these critical regions.

\subsection{Applications}
\label{sec:applications}

Our framework serves as a versatile engine for closed-loop autonomous driving simulation. Leveraging our disentangled dynamic representation, we enable flexible scene editing capabilities, such asvehicle removal and trajectory modification (see Fig.~9). These features facilitate the synthesis of diverse corner cases and clean background assets, providing a scalable and cost-effective solution for robust algorithm testing in digital twin environments.

\begin{table}[t]
    \centering
    \caption{Ablation study on different geometric priors for densification. Our strategy is compatible with various point sources. Combining LiDAR with 3D-VFM leverages both geometric accuracy and scene coverage, yielding the best reconstruction quality. Best results are \textbf{bolded}.}
    \label{tab:ablation_candidates}
    \begin{tabular}{l c c c}
        \toprule
        Geometric Prior & PSNR $\uparrow$ & SSIM $\uparrow$ & LPIPS $\downarrow$ \\
        \midrule
        3D-VFM & 24.04 & 0.700 & 0.340 \\
        LiDAR & 24.14 & 0.703 & 0.335 \\
        \midrule
        \textbf{LiDAR + 3D-VFM} & \textbf{24.19} & \textbf{0.705} & \textbf{0.329} \\
        \bottomrule
    \end{tabular}
\end{table}

\section{Conclusion}
\label{sec:conclusion}
In this paper, we presented L2D2-GS, a unified framework for the generalizable reconstruction of dynamic urban environments. Addressing the scalability bottlenecks and multi-view inconsistencies inherent in existing feedforward prediction models, we reformulated generalizable reconstruction as a robust, iterative process of optimization and densification. To overcome the lack of explicit supervision for primitive growth, we introduced a fully self-supervised densification policy that dynamically allocates Gaussian primitives by deriving explicit reward signals from global reconstruction gains. Furthermore, to ensure structural stability during the highly volatile early stages of optimization, we proposed a geometric regularization mechanism utilizing reparameterization to constrain the optimization manifold, effectively preventing primitives from collapsing into unfavorable local optima. 

Extensive evaluations on large-scale datasets demonstrated that our method achieves state-of-the-art reconstruction fidelity and robust zero-shot generalization. We believe this iterative, learning-to-densify paradigm bridges the gap between fast feedforward inference and high-quality per-scene optimization. In future work, we plan to extend this framework by incorporating temporal consistency regularizations across longer sequences and exploring its potential for real-time, closed-loop autonomous driving simulation.




\bibliographystyle{IEEEtran}
\bibliography{main}

@String(AAAI  = {AAAI})

@String(TOG   = {ACM Trans. Graph.})

@String(TOG   = {ACM TOG})

@inproceedings{nam2025generative,
  title={Generative Densification: Learning to Densify Gaussians for High-Fidelity Generalizable 3D Reconstruction},
  author={Nam, Seungtae and Sun, Xiangyu and Kang, Gyeongjin and Lee, Younggeun and Oh, Seungjun and Park, Eunbyung},
  booktitle={Proceedings of the Computer Vision and Pattern Recognition Conference},
  pages={26683--26693},
  year={2025}
}

@article{weng2025gaussianlens,
  title={GaussianLens: Localized High-Resolution Reconstruction via On-Demand Gaussian Densification},
  author={Weng, Yijia and Wang, Zhicheng and Peng, Songyou and Xie, Saining and Zhou, Howard and Guibas, Leonidas J},
  journal={arXiv preprint arXiv:2509.25603},
  year={2025}
}

@article{liu2025quicksplat,
  title={QuickSplat: Fast 3D Surface Reconstruction via Learned Gaussian Initialization},
  author={Liu, Yueh-Cheng and H{\"o}llein, Lukas and Nie{\ss}ner, Matthias and Dai, Angela},
  journal={arXiv preprint arXiv:2505.05591},
  year={2025}
}

@inproceedings{chan2024point,
  title={Point cloud densification for 3d gaussian splatting from sparse input views},
  author={Chan, Kin-Chung and Xiao, Jun and Goshu, Hana Lebeta and Lam, Kin-Man},
  booktitle={Proceedings of the 32nd ACM International Conference on Multimedia},
  pages={896--904},
  year={2024}
}

@inproceedings{patle2025ad,
  title={AD-GS: Alternating Densification for Sparse-Input 3D Gaussian Splatting},
  author={Patle, Gurutva and Girgaonkar, Nilay and Somraj, Nagabhushan and Soundararajan, Rajiv},
  booktitle={Proceedings of the SIGGRAPH Asia 2025 Conference Papers},
  pages={1--11},
  year={2025}
}

@inproceedings{ye2024absgs,
  title={Absgs: Recovering fine details in 3d gaussian splatting},
  author={Ye, Zongxin and Li, Wenyu and Liu, Sidun and Qiao, Peng and Dou, Yong},
  booktitle={Proceedings of the 32nd ACM International Conference on Multimedia},
  pages={1053--1061},
  year={2024}
}

@inproceedings{kim2024color,
  title={Color-cued efficient densification method for 3d gaussian splatting},
  author={Kim, Sieun and Lee, Kyungjin and Lee, Youngki},
  booktitle={Proceedings of the IEEE/CVF Conference on Computer Vision and Pattern Recognition},
  pages={775--783},
  year={2024}
}

@inproceedings{rota2024revising,
  title={Revising densification in gaussian splatting},
  author={Rota Bul{\`o}, Samuel and Porzi, Lorenzo and Kontschieder, Peter},
  booktitle={European Conference on Computer Vision},
  pages={347--362},
  year={2024},
  organization={Springer}
}

@inproceedings{mohamad2025denser,
  title={DENSER: 3D Gaussian Splatting for Scene Reconstruction of Dynamic Urban Environments},
  author={Mohamad, Mahmud A and Elghazaly, Gamal and Hubert, Arthur and Frank, Raphael},
  booktitle={2025 IEEE International Conference on Robotics and Automation (ICRA)},
  pages={2701--2707},
  year={2025},
  organization={IEEE}
}

@article{kerbl20233d,
  title={3D Gaussian splatting for real-time radiance field rendering.},
  author={Kerbl, Bernhard and Kopanas, Georgios and Leimk{\"u}hler, Thomas and Drettakis, George},
  journal={ACM Trans. Graph.},
  volume={42},
  number={4},
  pages={139--1},
  year={2023}
}

@inproceedings{miao2025evolsplat,
  title={Evolsplat: Efficient volume-based gaussian splatting for urban view synthesis},
  author={Miao, Sheng and Huang, Jiaxin and Bai, Dongfeng and Yan, Xu and Zhou, Hongyu and Wang, Yue and Liu, Bingbing and Geiger, Andreas and Liao, Yiyi},
  booktitle={Proceedings of the Computer Vision and Pattern Recognition Conference},
  pages={11286--11296},
  year={2025}
}

@inproceedings{wei2025omni,
  title={Omni-scene: Omni-gaussian representation for ego-centric sparse-view scene reconstruction},
  author={Wei, Dongxu and Li, Zhiqi and Liu, Peidong},
  booktitle={Proceedings of the Computer Vision and Pattern Recognition Conference},
  pages={22317--22327},
  year={2025}
}

@inproceedings{xu2025depthsplat,
  title={Depthsplat: Connecting gaussian splatting and depth},
  author={Xu, Haofei and Peng, Songyou and Wang, Fangjinhua and Blum, Hermann and Barath, Daniel and Geiger, Andreas and Pollefeys, Marc},
  booktitle={Proceedings of the Computer Vision and Pattern Recognition Conference},
  pages={16453--16463},
  year={2025}
}

@inproceedings{chen2024mvsplat,
  title={Mvsplat: Efficient 3d gaussian splatting from sparse multi-view images},
  author={Chen, Yuedong and Xu, Haofei and Zheng, Chuanxia and Zhuang, Bohan and Pollefeys, Marc and Geiger, Andreas and Cham, Tat-Jen and Cai, Jianfei},
  booktitle={European Conference on Computer Vision},
  pages={370--386},
  year={2024},
  organization={Springer}
}

@inproceedings{charatan2024pixelsplat,
  title={pixelsplat: 3d gaussian splats from image pairs for scalable generalizable 3d reconstruction},
  author={Charatan, David and Li, Sizhe Lester and Tagliasacchi, Andrea and Sitzmann, Vincent},
  booktitle={Proceedings of the IEEE/CVF conference on computer vision and pattern recognition},
  pages={19457--19467},
  year={2024}
}

@article{ye2024no,
  title={No pose, no problem: Surprisingly simple 3d gaussian splats from sparse unposed images},
  author={Ye, Botao and Liu, Sifei and Xu, Haofei and Li, Xueting and Pollefeys, Marc and Yang, Ming-Hsuan and Peng, Songyou},
  journal={arXiv preprint arXiv:2410.24207},
  year={2024}
}

@article{jiang2025anysplat,
  title={Anysplat: Feed-forward 3d gaussian splatting from unconstrained views},
  author={Jiang, Lihan and Mao, Yucheng and Xu, Linning and Lu, Tao and Ren, Kerui and Jin, Yichen and Xu, Xudong and Yu, Mulin and Pang, Jiangmiao and Zhao, Feng and others},
  journal={ACM Transactions on Graphics (TOG)},
  volume={44},
  number={6},
  pages={1--16},
  year={2025},
  publisher={ACM New York, NY, USA}
}

@article{smart2024splatt3r,
  title={Splatt3r: Zero-shot gaussian splatting from uncalibrated image pairs},
  author={Smart, Brandon and Zheng, Chuanxia and Laina, Iro and Prisacariu, Victor Adrian},
  journal={arXiv preprint arXiv:2408.13912},
  year={2024}
}

@inproceedings{zhang2025transplat,
  title={Transplat: Generalizable 3d gaussian splatting from sparse multi-view images with transformers},
  author={Zhang, Chuanrui and Zou, Yingshuang and Li, Zhuoling and Yi, Minmin and Wang, Haoqian},
  booktitle={Proceedings of the AAAI Conference on Artificial Intelligence},
  volume={39},
  pages={9869--9877},
  year={2025}
}

@article{chen2024mvsplat360,
  title={Mvsplat360: Feed-forward 360 scene synthesis from sparse views},
  author={Chen, Yuedong and Zheng, Chuanxia and Xu, Haofei and Zhuang, Bohan and Vedaldi, Andrea and Cham, Tat-Jen and Cai, Jianfei},
  journal={Advances in Neural Information Processing Systems},
  volume={37},
  pages={107064--107086},
  year={2024}
}

@inproceedings{zhang2024gs,
  title={Gs-lrm: Large reconstruction model for 3d gaussian splatting},
  author={Zhang, Kai and Bi, Sai and Tan, Hao and Xiangli, Yuanbo and Zhao, Nanxuan and Sunkavalli, Kalyan and Xu, Zexiang},
  booktitle={European Conference on Computer Vision},
  pages={1--19},
  year={2024},
  organization={Springer}
}

@inproceedings{xu2024grm,
  title={Grm: Large gaussian reconstruction model for efficient 3d reconstruction and generation},
  author={Xu, Yinghao and Shi, Zifan and Yifan, Wang and Chen, Hansheng and Yang, Ceyuan and Peng, Sida and Shen, Yujun and Wetzstein, Gordon},
  booktitle={European Conference on Computer Vision},
  pages={1--20},
  year={2024},
  organization={Springer}
}

@inproceedings{ziwen2025long,
  title={Long-lrm: Long-sequence large reconstruction model for wide-coverage gaussian splats},
  author={Ziwen, Chen and Tan, Hao and Zhang, Kai and Bi, Sai and Luan, Fujun and Hong, Yicong and Fuxin, Li and Xu, Zexiang},
  booktitle={Proceedings of the IEEE/CVF International Conference on Computer Vision},
  pages={4349--4359},
  year={2025}
}

@inproceedings{szymanowicz2024splatter,
  title={Splatter image: Ultra-fast single-view 3d reconstruction},
  author={Szymanowicz, Stanislaw and Rupprecht, Chrisitian and Vedaldi, Andrea},
  booktitle={Proceedings of the IEEE/CVF conference on computer vision and pattern recognition},
  pages={10208--10217},
  year={2024}
}

@inproceedings{chen2024g3r,
  title={G3r: Gradient guided generalizable reconstruction},
  author={Chen, Yun and Wang, Jingkang and Yang, Ze and Manivasagam, Sivabalan and Urtasun, Raquel},
  booktitle={European Conference on Computer Vision},
  pages={305--323},
  year={2024},
  organization={Springer}
}

@article{xu2025resplat,
  title={Resplat: Learning recurrent gaussian splats},
  author={Xu, Haofei and Barath, Daniel and Geiger, Andreas and Pollefeys, Marc},
  journal={arXiv preprint arXiv:2510.08575},
  year={2025}
}

@article{wang2025flux4d,
  title={Flux4d: Flow-based unsupervised 4d reconstruction},
  author={Wang, Jingkang and Che, Henry and Chen, Yun and Yang, Ze and Goli, Lily and Manivasagam, Sivabalan and Urtasun, Raquel},
  journal={arXiv preprint arXiv:2512.03210},
  year={2025}
}

@article{chen2024splatformer,
  title={Splatformer: Point transformer for robust 3d gaussian splatting},
  author={Chen, Yutong and Mihajlovic, Marko and Chen, Xiyi and Wang, Yiming and Prokudin, Sergey and Tang, Siyu},
  journal={arXiv preprint arXiv:2411.06390},
  year={2024}
}

@inproceedings{hess2025splatad,
  title={Splatad: Real-time lidar and camera rendering with 3d gaussian splatting for autonomous driving},
  author={Hess, Georg and Lindstr{\"o}m, Carl and Fatemi, Maryam and Petersson, Christoffer and Svensson, Lennart},
  booktitle={Proceedings of the Computer Vision and Pattern Recognition Conference},
  pages={11982--11992},
  year={2025}
}

@inproceedings{yan2024street,
  title={Street gaussians: Modeling dynamic urban scenes with gaussian splatting},
  author={Yan, Yunzhi and Lin, Haotong and Zhou, Chenxu and Wang, Weijie and Sun, Haiyang and Zhan, Kun and Lang, Xianpeng and Zhou, Xiaowei and Peng, Sida},
  booktitle={European Conference on Computer Vision},
  pages={156--173},
  year={2024},
  organization={Springer}
}

@article{chen2024omnire,
  title={Omnire: Omni urban scene reconstruction},
  author={Chen, Ziyu and Yang, Jiawei and Huang, Jiahui and de Lutio, Riccardo and Esturo, Janick Martinez and Ivanovic, Boris and Litany, Or and Gojcic, Zan and Fidler, Sanja and Pavone, Marco and others},
  journal={arXiv preprint arXiv:2408.16760},
  year={2024}
}

@inproceedings{peng2025desire,
  title={Desire-gs: 4d street gaussians for static-dynamic decomposition and surface reconstruction for urban driving scenes},
  author={Peng, Chensheng and Zhang, Chengwei and Wang, Yixiao and Xu, Chenfeng and Xie, Yichen and Zheng, Wenzhao and Keutzer, Kurt and Tomizuka, Masayoshi and Zhan, Wei},
  booktitle={Proceedings of the Computer Vision and Pattern Recognition Conference},
  pages={6782--6791},
  year={2025}
}

@inproceedings{xu2025ad,
  title={AD-GS: Object-aware B-Spline Gaussian splatting for self-supervised autonomous driving},
  author={Xu, Jiawei and Deng, Kai and Fan, Zexin and Wang, Shenlong and Xie, Jin and Yang, Jian},
  booktitle={Proceedings of the IEEE/CVF International Conference on Computer Vision},
  pages={24770--24779},
  year={2025}
}

@inproceedings{tang2023torchsparse++,
  title={Torchsparse++: Efficient training and inference framework for sparse convolution on gpus},
  author={Tang, Haotian and Yang, Shang and Liu, Zhijian and Hong, Ke and Yu, Zhongming and Li, Xiuyu and Dai, Guohao and Wang, Yu and Han, Song},
  booktitle={Proceedings of the 56th Annual IEEE/ACM International Symposium on Microarchitecture},
  pages={225--239},
  year={2023}
}

@article{yang2024storm,
  title={Storm: Spatio-temporal reconstruction model for large-scale outdoor scenes},
  author={Yang, Jiawei and Huang, Jiahui and Chen, Yuxiao and Wang, Yan and Li, Boyi and You, Yurong and Sharma, Apoorva and Igl, Maximilian and Karkus, Peter and Xu, Danfei and others},
  journal={arXiv preprint arXiv:2501.00602},
  year={2024}
}

@inproceedings{xiao2021pandaset,
  title={Pandaset: Advanced sensor suite dataset for autonomous driving},
  author={Xiao, Pengchuan and Shao, Zhenlei and Hao, Steven and Zhang, Zishuo and Chai, Xiaolin and Jiao, Judy and Li, Zesong and Wu, Jian and Sun, Kai and Jiang, Kun and others},
  booktitle={2021 IEEE international intelligent transportation systems conference (ITSC)},
  pages={3095--3101},
  year={2021},
  organization={IEEE}
}

@inproceedings{sun2020scalability,
  title={Scalability in perception for autonomous driving: Waymo open dataset},
  author={Sun, Pei and Kretzschmar, Henrik and Dotiwalla, Xerxes and Chouard, Aurelien and Patnaik, Vijaysai and Tsui, Paul and Guo, James and Zhou, Yin and Chai, Yuning and Caine, Benjamin and others},
  booktitle={Proceedings of the IEEE/CVF conference on computer vision and pattern recognition},
  pages={2446--2454},
  year={2020}
}

@article{keetha2025mapanything,
  title={Mapanything: Universal feed-forward metric 3d reconstruction},
  author={Keetha, Nikhil and M{\"u}ller, Norman and Sch{\"o}nberger, Johannes and Porzi, Lorenzo and Zhang, Yuchen and Fischer, Tobias and Knapitsch, Arno and Zauss, Duncan and Weber, Ethan and Antunes, Nelson and others},
  journal={arXiv preprint arXiv:2509.13414},
  year={2025}
}

@inproceedings{wang2025vggt,
  title={Vggt: Visual geometry grounded transformer},
  author={Wang, Jianyuan and Chen, Minghao and Karaev, Nikita and Vedaldi, Andrea and Rupprecht, Christian and Novotny, David},
  booktitle={Proceedings of the Computer Vision and Pattern Recognition Conference},
  pages={5294--5306},
  year={2025}
}

@article{chen2025dggt,
  title={DGGT: Feedforward 4D Reconstruction of Dynamic Driving Scenes using Unposed Images},
  author={Chen, Xiaoxue and Xiong, Ziyi and Chen, Yuantao and Li, Gen and Wang, Nan and Luo, Hongcheng and Chen, Long and Sun, Haiyang and Wang, Bing and Chen, Guang and others},
  journal={arXiv preprint arXiv:2512.03004},
  year={2025}
}

@article{yang2023emernerf,
  title={Emernerf: Emergent spatial-temporal scene decomposition via self-supervision},
  author={Yang, Jiawei and Ivanovic, Boris and Litany, Or and Weng, Xinshuo and Kim, Seung Wook and Li, Boyi and Che, Tong and Xu, Danfei and Fidler, Sanja and Pavone, Marco and others},
  journal={arXiv preprint arXiv:2311.02077},
  year={2023}
}

@ARTICLE{Liang20264DGStream,
  author={Liang, Zhicheng and Zhang, Dayou and Shen, Linfeng and Zhang, Miao and Zhang, Jian and Ju, Bin and Dasari, Mallesham and Wang, Fangxin and Liu, Jiangchuan},
  journal={IEEE Transactions on Multimedia}, 
  title={4DGStream: Variable Bitrate Dynamic Gaussian Splatting Streaming}, 
  year={2026},
  volume={},
  number={},
  pages={1-15},
  doi={10.1109/TMM.2026.3668629}}

@ARTICLE{Duan2026BUGS,
  author={Duan, Fan and Zhang, Yumeng and Li, Xiaofan and Tan, Xiao and Wang, Jingdong and Chen, Li},
  journal={IEEE Transactions on Multimedia}, 
  title={BUGS: Universal 3D Gaussian Splatting with a Bi-directional Gaussian Growing Mechanism}, 
  year={2026},
  volume={},
  number={},
  pages={1-11},
  doi={10.1109/TMM.2026.3668589}}

@ARTICLE{Shuai2026Adversarial,
  author={Shuai, Hui and Shi, Yucheng and Sun, Yubao and Liu, Qingshan},
  journal={IEEE Transactions on Multimedia}, 
  title={Adversarial Pruning Networks for Compact 3D Gaussian Splatting}, 
  year={2026},
  volume={28},
  number={},
  pages={1080-1089},
  doi={10.1109/TMM.2025.3632681}}

\clearpage

\ifmain
\else
\begin{figure*}
    \centering
    \includegraphics[width=0.8\linewidth]{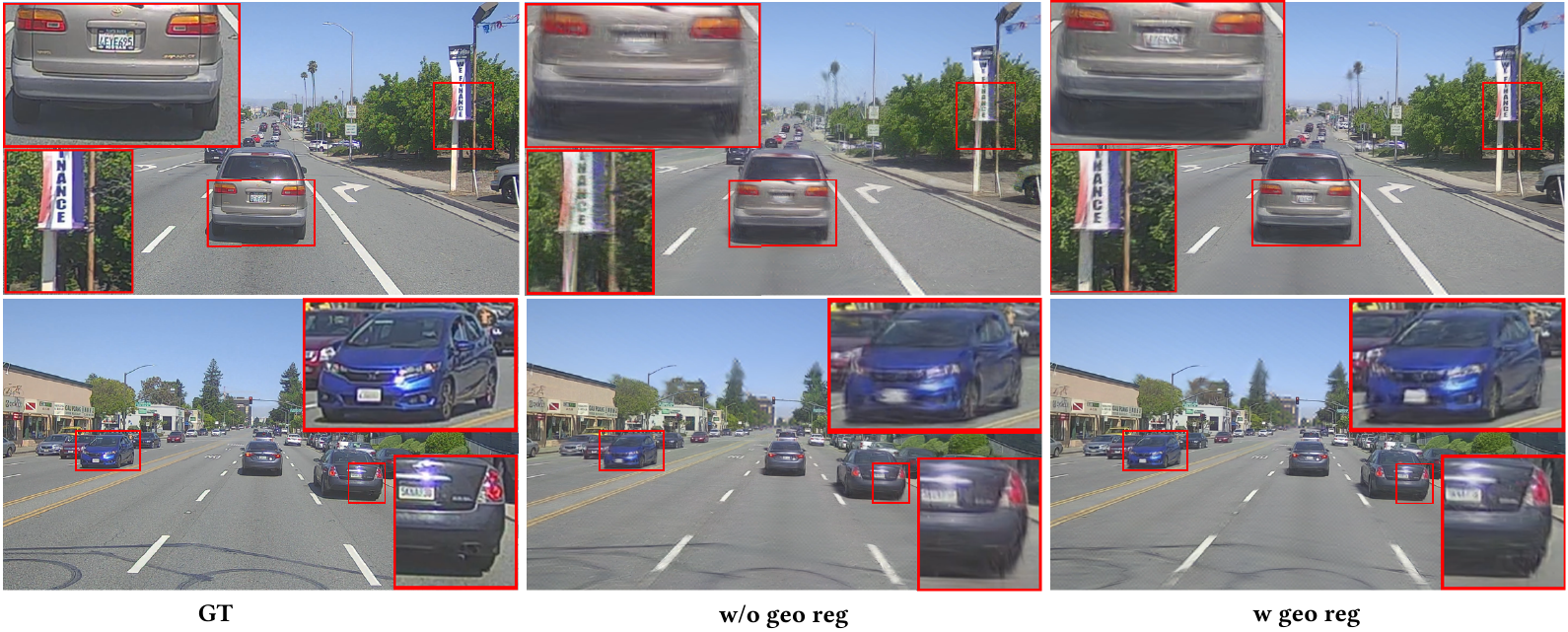}
    \caption{\textbf{Effectiveness of geometric regularization.} 
    Without reparameterization, the reconstruction is plagued by degenerate artifacts (highlighted in red boxes). Our method effectively suppresses these irregularities, ensuring robust geometric convergence.}
    \label{fig:geo_reg}
\end{figure*}

\begin{figure*}
    \centering
    \includegraphics[width=1\linewidth]{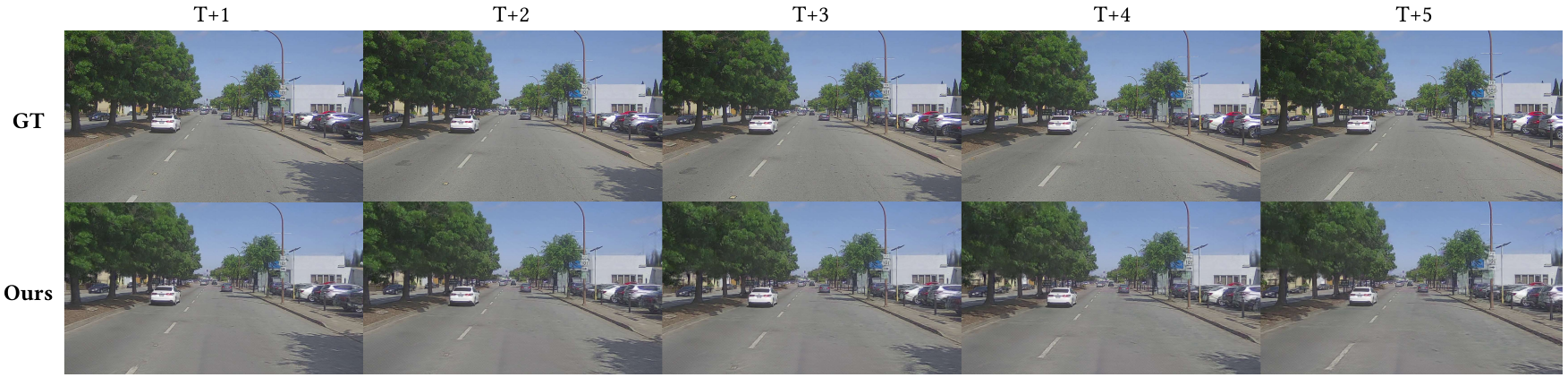}
    \caption{\textbf{Future prediction under novel views.} Our framework generalizes to unseen timestamps, rendering high-fidelity images of future states while allowing for free viewpoint navigation.}
    \label{fig:extrapolation}
\end{figure*}

\section{Additional Experimental Results}

\noindent\textbf{Experimental Results in G3R Configuration~\cite{chen2024g3r}.}
To ensure a comprehensive and fair comparison, we further evaluated our method following the exact dataset split and experimental protocol defined in G3R~\cite{chen2024g3r}. We trained our model from scratch under this setting and conducted testing on the designated test scenes (Scenes 1, 30, 40, 80, 90, 110, and 120). The quantitative results are reported in Table~\ref{tab:g3r_split}, demonstrating that our method consistently outperforms the baseline across all metrics.

\vspace{1pt}
\noindent\textbf{Effect of Geometry Reparameterization.}
As illustrated in Fig.~\ref{fig:geo_reg}, our reparameterization strategy significantly enhances reconstruction quality by effectively mitigating geometric degradation in the foreground. While the baseline suffers from needle-like artifacts, our method maintains clean and coherent geometry in these critical regions.

\vspace{1pt}
\noindent\textbf{Future Extrapolation.}
We evaluate its performance on extrapolating to future frames beyond the training horizon. As illustrated in Fig. \ref{fig:extrapolation}, our method generates plausible and temporally consistent views for these unseen frames. This demonstrates that our model has learned a generalized scene representation rather than merely overfitting to the observed timestamps.

\vspace{1pt}
\noindent\textbf{Scene Editting Application.}
Our framework serves as a versatile engine for closed-loop autonomous driving simulation. Leveraging our disentangled dynamic representation, we enable flexible scene editing capabilities, such asvehicle removal and trajectory modification (see Fig.~\ref{fig:application}). These features facilitate the synthesis of diverse corner cases and clean background assets, providing a scalable and cost-effective solution for robust algorithm testing in digital twin environments.

\begin{figure*}[t]
    \centering
    \includegraphics[width=\linewidth]{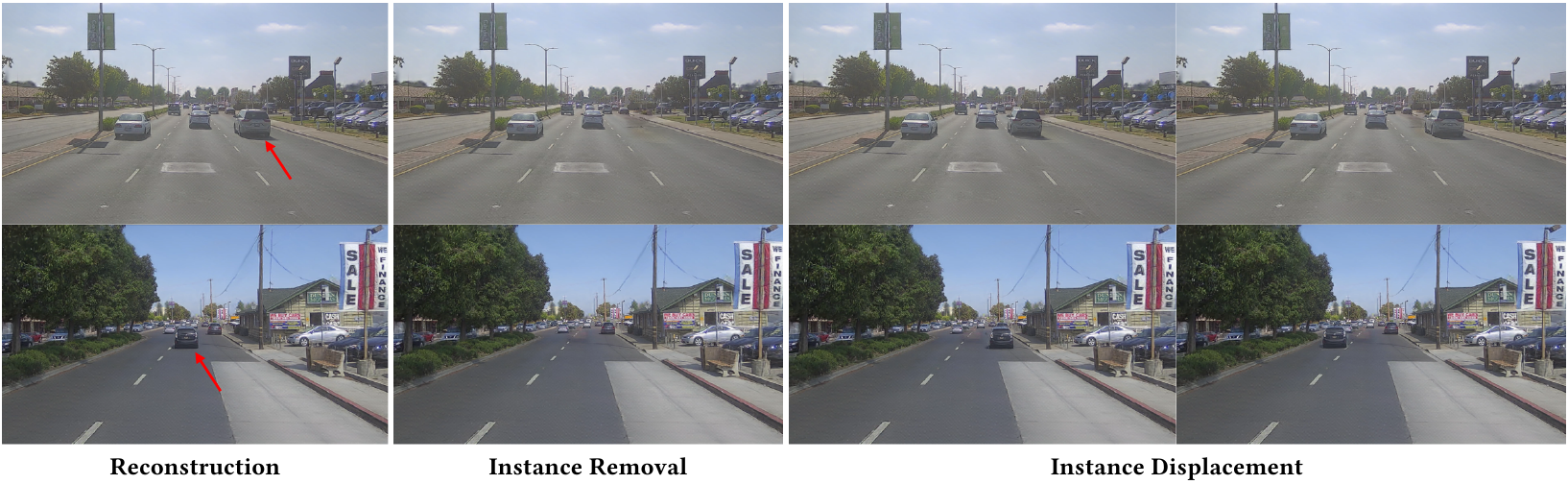}
    \caption{\textbf{Scene editing applications.} Our disentangled representation enables flexible manipulation of dynamic agents, such as vehicle removal and trajectory modification. This facilitates the synthesis of diverse corner cases for robust autonomous driving simulation.}
    \label{fig:application}
\end{figure*}

\begin{figure*}[t]
    \centering
    \includegraphics[width=1\linewidth]{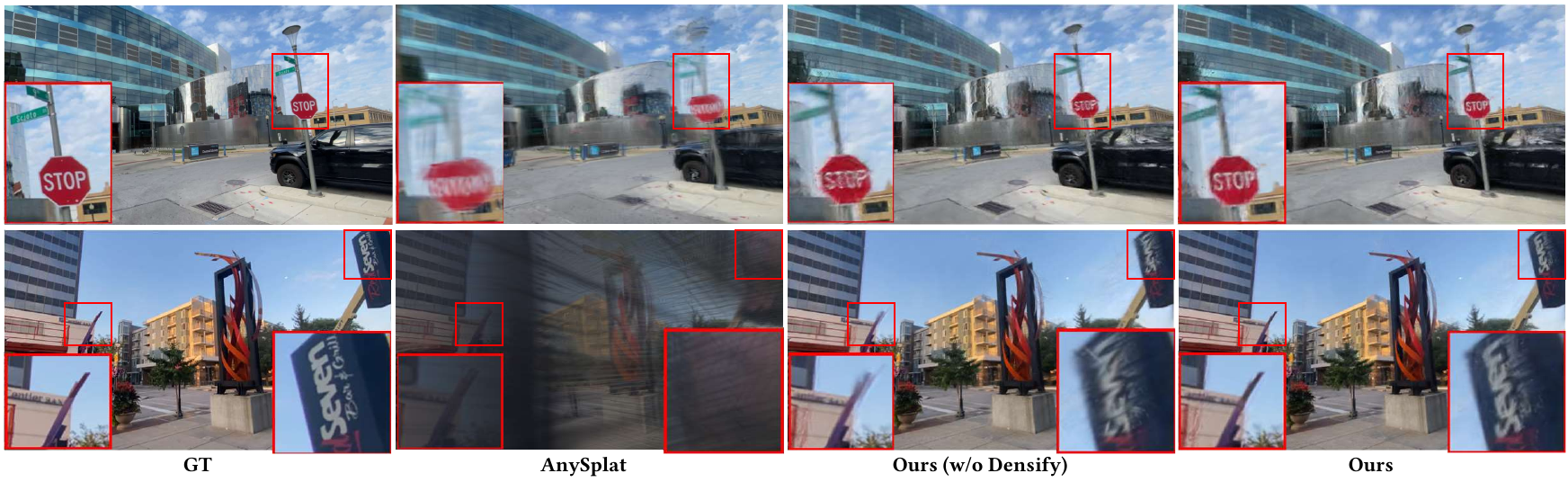}
    \caption{Qualitative comparison on the DL3DV dataset. We use a 64-input-view setting to evaluate multiview consistency. AnySplat struggles with dense view input, whereas our method remains robust. The last two columns highlight the visual improvements brought by our densification module.}
    \label{fig:dl3dv_qualitative}
\end{figure*}


\begin{table}[t]
    \centering
    \caption{\textbf{Quantitative comparison following the G3R protocol.} We report the performance on full sequence interpolation using the dataset split defined in G3R~\cite{chen2024g3r}.}
    \label{tab:g3r_split}
    \begin{tabular}{l|ccc}
    \toprule
    Method & PSNR $\uparrow$ & SSIM $\uparrow$ & LPIPS $\downarrow$ \\
    \midrule
    G3R & 25.22 & 0.742 & 0.371 \\
    \textbf{Ours} & \textbf{26.87} & \textbf{0.789} & \textbf{0.254} \\
    \bottomrule
    \end{tabular}
\end{table}

\begin{table}[htbp]
    \centering
    \caption{Ablation study on the number of nearest neighbors ($N_k$), query points ($N_q$), and densified points ($N_d$). The metrics report the relative performance changes ($\Delta$) compared to the baseline configuration.}
    \label{tab:hyper_ablation}
    \setlength{\tabcolsep}{4pt} 
    \begin{tabular}{ccc|ccc}
        \toprule
        $N_k$ & $N_q (\times 10^6)$ & $N_d (\times 10^5)$ & $\Delta$PSNR $\uparrow$ & $\Delta$SSIM $\uparrow$ & $\Delta$LPIPS $\downarrow$ \\
        \midrule
        16 & 4 & 4 & \textbf{0.00} & \textbf{\phantom{-}0.000} & \textbf{0.000} \\
        16 & 2 & 4 & -0.08 & -0.003 & 0.003 \\
        16 & 1 & 4 & -0.18 & -0.005 & 0.005 \\
        16 & 4 & 2 & -0.01 & -0.001 & 0.004 \\
        16 & 4 & 1 & -0.06 & -0.003 & 0.008 \\
        8 & 4 & 4 & -0.08 & -0.013 & 0.003 \\
        \bottomrule
    \end{tabular}
\end{table}

\vspace{1pt}
\noindent\textbf{Hyperparameter Sensitivity Analysis.}
We investigate the impact of different densification configurations on the objective reconstruction performance, as reported in Table~\ref{tab:hyper_ablation}. The results indicate that reducing the number of candidate points degrades the rendering quality, whereas decreasing the number of actually densified points yields a relatively marginal performance drop. This observation validates the efficacy of our proposed strategy, demonstrating its ability to accurately identify and select the most crucial points from the candidate set for densification.

Furthermore, we evaluate the sensitivity of our model to the number of nearest neighbors, $N_k$, in the KNN attention mechanism. We retrain the densification network from scratch with the modified hyperparameter. As shown in the table, decreasing $N_k$ harms overall performance, indicating that a sufficient local receptive field is necessary for robust feature aggregation and accurate densification policy learning.

\vspace{1pt}
\noindent\textbf{Generalization Experiment on DL3DV.}
To ensure a fair comparison with feedforward Gaussian prediction methods and to validate the generalization capabilities of our approach, we finetuned our model on the DL3DV dataset and evaluated the performance of both our overall framework and the densification network. Specifically, we selected a subset of 1,000 scenes from the DL3DV dataset at a resolution of $960 \times 540$, allocating 900 scenes for training and 100 scenes for testing. Because the DL3DV dataset do not contain Lidar, we only use MapAnything~\cite{keetha2025mapanything} as geometry prior.

As reported in Table \ref{tab:dl3dv_generalization}, our method achieves substantially better performance than the representative feedforward Gaussian prediction method (AnySplat). This substantial improvement is primarily attributed to our joint iterative optimization and densification framework, which more effectively handles multi-view consistency across dense viewpoints. Furthermore, conducting an ablation by disabling the densification module (``w/o densify'') results in a noticeable performance drop, demonstrating the efficacy of our proposed densification network. 

\begin{table}[t]
  \centering
  \caption{NVS result on the DL3DV dataset. We evaluate the generalization capability of our framework against the SoTA feedforward method (AnySplat) and an ablation of our densification module. The best results are \textbf{bold}.}
  \label{tab:dl3dv_generalization}
  \setlength{\tabcolsep}{4pt} 
  \begin{tabular}{clcccc}
    \toprule
    Input Views & Method & PSNR $\uparrow$ & SSIM $\uparrow$ & LPIPS $\downarrow$ & $N_{gs}$ \\
    \midrule
    \multirow{3}{*}{32} 
    & AnySplat & 18.76 & 0.569 & 0.389 & 5M \\
    & Ours (w/o densify) & 23.97 & 0.737 & 0.256 & 700K \\
    & Ours & \textbf{24.19} & \textbf{0.746} & \textbf{0.249} & 700K \\
    \midrule
    \multirow{3}{*}{64} 
    & AnySplat & 17.33 & 0.518 & 0.458 & 10M \\
    & Ours (w/o densify) & 23.63 & 0.721 & 0.277 & 700K \\
    & Ours & \textbf{23.88} & \textbf{0.733} & \textbf{0.269} & 700K \\
    \bottomrule
  \end{tabular}
\end{table}

Qualitative comparisons are provided in Fig. \ref{fig:dl3dv_qualitative}. As illustrated, AnySplat struggles to maintain rendering fidelity when provided with dense input views. Furthermore, we observe that AnySplat occasionally fails under inward-facing camera configurations (bottom row), which is likely attributable to inaccurate geometry estimation. In contrast, our approach exhibits strong robustness in these challenging scenarios. Additionally, the visual comparisons in the last two columns clearly demonstrate the effectiveness of our proposed densification method in capturing fine details.

\noindent\textbf{Scene Reconstruction Trajectory.}
Our scene reconstruction module demonstrates the capability to consistently optimize the scene representations over time. To explicitly illustrate this progression, Fig.~\ref{fig:opt_curve} presents the evolution of quantitative metrics across different reconstruction timesteps, with the results averaged over the test set. The curves exhibit a stable and continuous improvement in rendering fidelity, confirming the effectiveness of our progressive optimization strategy.

Furthermore, we provide a qualitative visual trajectory of the reconstruction process in Fig.~\ref{fig:opt_traj}. As observed in the subjective results, our method rapidly establishes the coarse global geometry and overall topological structure of the scene within merely $4$ timesteps. The subsequent optimization steps are primarily dedicated to the meticulous refinement of high-frequency local details, such as the thin structures of street lamp poles and the legible text on traffic signs. This coarse-to-fine convergence behavior ensures both efficient structural initialization and high-fidelity detail preservation.

\begin{figure}[htbp]
    \centering
    \includegraphics[width=\linewidth]{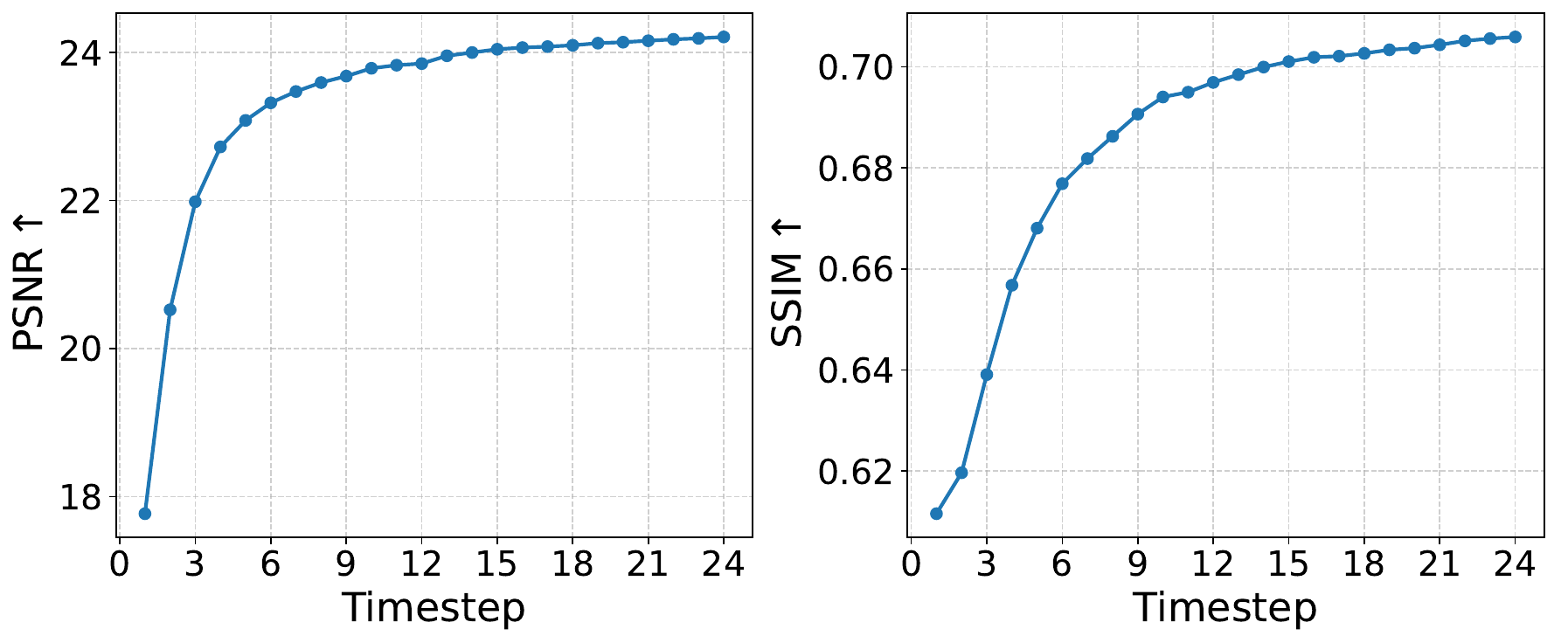}
    \caption{\textbf{Quantitative Reconstruction Trajectory.} Evolution of objective metrics across reconstruction timesteps, averaged over the test scenes. The consistent trends demonstrate the stable convergence of our reconstruction module.}
    \label{fig:opt_curve}
\end{figure}

\begin{figure}[htbp]
    \centering
    \includegraphics[width=\linewidth]{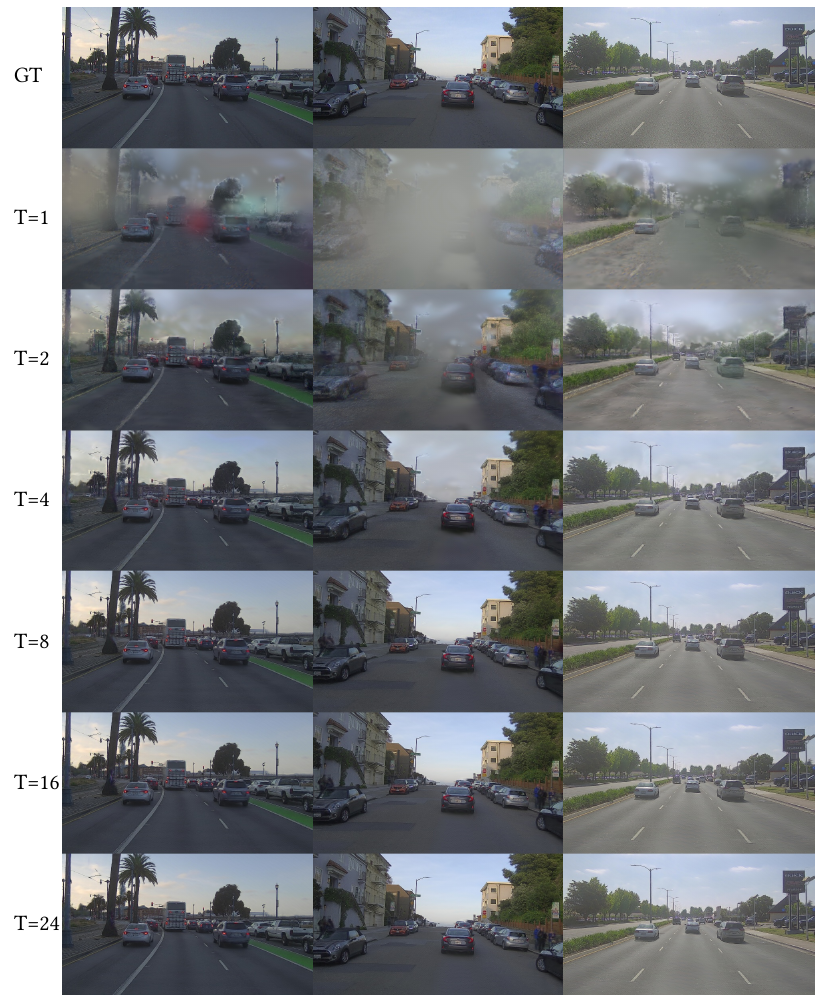}
    \caption{\textbf{Qualitative Reconstruction Trajectory.} Visualizations of the rendered novel views at different timesteps. Our method recovers the coarse scene structure efficiently within the first 4 timesteps, while the subsequent steps progressively refine high-frequency geometric and textural details (e.g., street lamps and textual signs).}
    \label{fig:opt_traj}
\end{figure}

\section{Implementation Details}

\vspace{1pt}
\noindent\textbf{Neural Gaussian Representation.}
We formulate the dynamic scene using a hybrid neural Gaussian representation. Specifically, each Gaussian primitive is defined as a tuple $\mathcal{G} = \{ \Theta, \mathbf{f} \}$ with a total of $46$ feature channels ($C=46$). Here, $\Theta$ denotes the $14$-dimensional explicit attributes (position, rotation, scaling, color, and opacity), while $\mathbf{f}$ represents a $32$-dimensional implicit latent feature. 


%
\vspace{1pt}
\noindent\textbf{Scene Initialization.}
We initialize the positions of the 3D neural Gaussians using downsampled 3D LiDAR points. We voxelize these points to generate voxel-wise Gaussians. To explicitly handle scene dynamics, we utilize Ground Truth (GT) cuboids to initialize the dynamic nodes, ensuring that dynamic Gaussians are correctly instantiated within the moving objects. For geometric attributes, the scale of each Gaussian is initialized isotropically as the distance to its third nearest neighbor to ensure sufficient coverage, while the rotation is set to identity and the opacity to 0.7. Regarding the neural features, unlike previous methods that use random initialization, we initialize all other feature channels to zero. We empirically find that this zero-initialization strategy facilitates more stable convergence during training stages.

\vspace{1pt}
\noindent\textbf{Metrics.}
Following standard evaluation protocols, we report Peak Signal-to-Noise Ratio (PSNR), Structural Similarity (SSIM) , and Learned Perceptual Image Patch Similarity (LPIPS) to quantitatively assess the photorealism of novel view synthesis. 
For LPIPS, we utilize the \textbf{AlexNet} backbone, as it is the standard configuration in recent view synthesis benchmarks. Additionally, to evaluate efficiency, we report the reconstruction time measured on a single NVIDIA H20 GPU.

\section{Baseline Implementation Details}

\noindent\textbf{Street Gaussians.}
To benchmark our method against SoTA dynamic scene reconstruction, we implement Street Gaussians \cite{yan2024street} using the codebase provided in the DriveStudio framework \footnote{\url{https://github.com/ziyc/drivestudio}}. We strictly adhere to the original training protocols and hyperparameter settings to ensure a fair evaluation. 
The scene is initialized with a total of 1 million primitives: this consists of 600K points downsampled from the aggregated LiDAR scans to provide a geometric coarse structure, augmented by 400K points uniformly sampled within the scene's bounding box. We train the model for 30,000 iterations per scene.

\noindent\textbf{OmniRe.}
We also utilize the official implementation of OmniRe \cite{chen2024omnire} provided within the DriveStudio framework for comparison. To ensure a consistent experimental setup, we adopt the same initialization strategy as Street Gaussians, employing a hybrid set of 600K downsampled LiDAR points and 400K randomly sampled points. The model is trained for 30,000 iterations per scene.
We disable the SMPL module as it utilizes additional human pose annotations. Moreover, the negligible pedestrian presence in the evaluation sequences renders specific articulation modeling redundant.

\noindent\textbf{AnySplat.}
We evaluate AnySplat \cite{jiang2025anysplat} using its official implementation \footnote{\url{https://github.com/InternRobotics/AnySplat}}.
For the testing configuration, we adopt a frame interpolation setting: the model takes two adjacent even frames ($I_{2k}, I_{2k+2}$) as input to synthesize the intermediate odd frame ($I_{2k+1}$). Note that the final odd frame in each sequence is excluded from evaluation due to the absence of a subsequent neighbor.
Regarding the model weights, we directly utilize the official pre-trained checkpoints, leveraging their generalization capability derived from large-scale training on diverse datasets, including outdoor scenes.
Since the architecture does not support high-resolution input, we downsample the input images to $448 \times 448$ for inference and bilinearly upsample the rendered outputs back to the original input resolution for metric calculation.

\noindent\textbf{G3R.}
To ensure a fair comparison, we faithfully reproduce the performance of G3R \cite{chen2024g3r} strictly following the implementation details provided in the original paper. The model is trained on 8 NVIDIA A100 GPUs for 50,000 iterations, which takes approximately two days to complete. We employ the AdamW optimizer with a learning rate of $10^{-4}$ and a weight decay of $10^{-5}$. Consistent with the original setting, we utilize a optimization process of 24 timesteps. This configuration corresponds to 2,000 scene iterations, ensuring that each training scene is processed approximately 20 times during optimization.

\begin{figure}[t]
    \centering
    \includegraphics[width=0.8\linewidth]{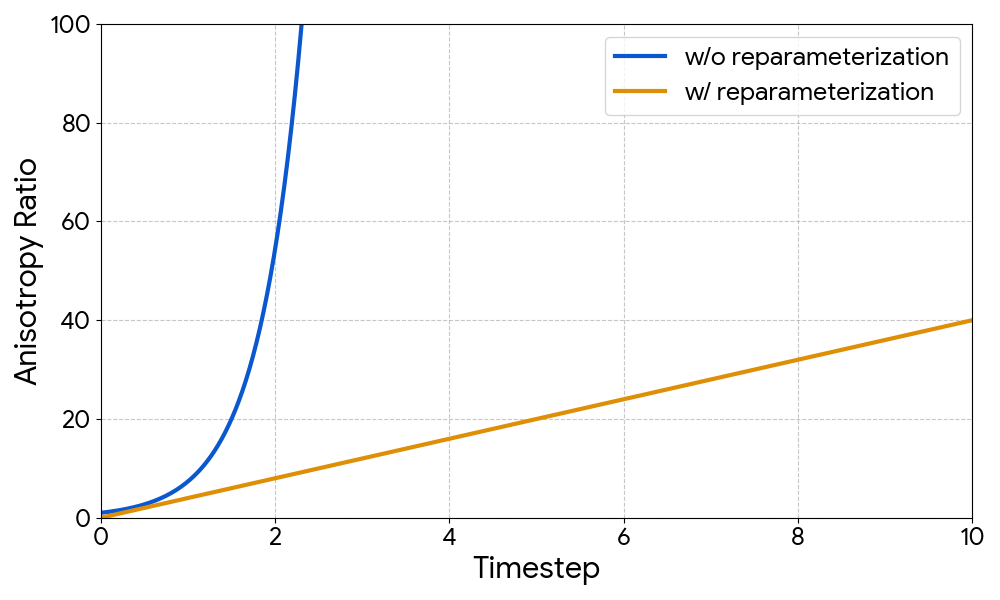}
    \caption{\textbf{Upper bound of reachable anisotropy ratio.} We plot the maximum anisotropy ratio theoretically reachable under cumulative updates (assuming a constant unit update step per iteration). Our reparameterization (orange) effectively constrains the optimization manifold, enforcing a linear upper bound on shape deformation. In contrast, the standard formulation (blue) allows the boundary to explode exponentially, making degenerate local optima easily accessible in early stages.}
    \label{fig:anisotropy_bound}
\end{figure}

\noindent\textbf{STORM.}
We evaluate STORM~\cite{yang2024storm} using the official open-source implementation\footnote{\url{https://github.com/NVlabs/GaussianSTORM}}. We strictly adhere to the testing protocols and hyperparameters described in the original paper, with the sole exception that our evaluation is conducted exclusively on the front-facing camera. To ensure a fair comparison with other baselines that operate at full resolution, we upsample the rendered images from STORM to the original video resolution via bilinear interpolation before calculating quantitative metrics.






\section{Analysis of Optimization Manifold Constraints}

To validate the theoretical effectiveness of our geometric regularization, we analyze the upper bound of the anisotropy ratio reachable by the optimization process. We define the anisotropy ratio as the quotient of the maximum scale to the minimum scale ($\lambda_{\max} / \lambda_{\min}$) of a Gaussian primitive. Fig.~\ref{fig:anisotropy_bound} visualizes the maximum possible anisotropy ratio allowed by the parameterization.
Specifically, we assume a normalized update step where the raw parameters controlling the scale (e.g., exponents or our latent factors) are updated within a range of $(-1, 1)$ at each timestep. The vertical axis represents the ratio between the maximum and minimum achievable scales ($\lambda_{\max} / \lambda_{\min}$) given these cumulative updates.

\begin{itemize}
    \item \textbf{Unconstrained Baseline (Blue):} Without regularization, the scale is typically modeled via an exponential activation ($\exp(s)$). In this space, a linear update in the exponent leads to an exponential divergence in the actual scale. Consequently, the maximum possible anisotropy ratio explodes exponentially, allowing the optimizer to easily reach degenerate, needle-like configurations ($>100$) within the first few iterations.
    
    \item \textbf{Ours with Reparameterization (Orange):} Our method reparameterizes the scale factors to map linear updates to a linear growth in anisotropy. As shown by the orange curve, the maximum reachable anisotropy increases linearly and slowly. This imposes a strict \textbf{structural constraint on the optimization manifold}: even if the gradients are large, the network is physically incapable of generating extreme needle-like artifacts in the early stages. This ''speed limit" on shape deformation forces the optimization to prioritize low-frequency geometric placement before refining high-frequency anisotropic details.
\end{itemize}

\fi

\vfill

\end{document}